\documentclass[a4paper, 10pt, conference]{ieeeconf}      % Use this line for a4
\usepackage{FG2024}

\usepackage{graphicx}
\usepackage{hyperref}
\usepackage{subcaption}
\usepackage{algorithm}
\usepackage{algorithmic}

\usepackage{multirow}
\usepackage{siunitx}
\usepackage{pifont} % check mark and cross mark
\usepackage{amsmath}
\usepackage{amssymb}
\usepackage{booktabs}

\usepackage{newfloat}
\usepackage{listings}

\FGfinalcopy % *** Uncomment this line for the final submission

\IEEEoverridecommandlockouts                              % This command is only
\def\FGPaperID{174}% *** Enter the FG2024 Paper ID here

\title{\LARGE \bf
3D Face Modeling via Weakly-supervised Disentanglement Network joint Identity-consistency Prior
}

\author{\parbox{16cm}{\centering
    {\large Guohao Li$^1$, Hongyu Yang$^{2,3}$, Di Huang$^1$ and Yunhong Wang$^1$}\\
    {\normalsize
    $^1$ School of Computer Science and Engineering, Beihang University, Beijing, China\\
    $^2$ Institute of Artificial Intelligence, Beihang University, Beijing, China\\
    $^3$ Shanghai Artificial Intelligence Laboratory, Shanghai, China\\
    \{liguohao, hongyuyang, dhuang, yhwang\}@buaa.edu.cn
    }}
    \thanks{}% <-this % stops a space
}

\newcommand{\ShockingName}{WSDF}
\newcommand{\MixerName}{Re-coupler}
\newcommand{\NBankName}{Neutral Bank}
\newcommand{\JLossName}{Jacobian Loss}

\newcommand{\ie}{\textit{i.e.}}  % that is (follows a comma)
\newcommand{\double}[1]{\mathbb{#1}}
\newcommand{\script}[1]{\mathcal{#1}}

\newcommand{\setof}[1]{\{#1\}}
\newcommand{\setelement}[2][i]{{#2}_{#1}}

\newcommand{\argmin}[1]{ \underset{#1}{\arg\min}}
\newcommand{\norm}[1]{\lVert#1\rVert}
\newcommand{\veczero}{\vec{0}}

\newcommand{\lowb}{$\downarrow$}  % lower is better

\newcommand*{\rpfloat}[2][2]{ {\num[mode=text,round-mode=places,round-precision=#1]{#2}} }

\newcommand*{\xrec}[1][]{X^r_#1}
\newcommand*{\xneu}[1][]{X^n_#1}

\newcommand{\namei}{G_{id}}            % notation of grouped
\newcommand{\nameg}{{id}}              % notation of un group
\newcommand{\disti}[1][]{q^{\scriptstyle id}_#1}
\newcommand{\diste}[1][]{q^{\scriptstyle exp}_#1}
\newcommand{\distg}[1][]{q^{\scriptstyle \nameg{}}_#1}
\newcommand{\codei}{z^{\scriptscriptstyle id}}
\newcommand{\codee}{z^{\scriptscriptstyle exp}}

\newcommand{\distnormal}{\boldsymbol{N}}
\newcommand{\distuniform}{\boldsymbol{U}}
\newcommand{\cmark}{\ding{51}}
\newcommand{\xmark}{\ding{55}}

\usepackage{fancyhdr}
\begin{document}

\ifFGfinal
\thispagestyle{empty}
\pagestyle{empty}
\else
\author{Anonymous FG2024 submission\\ Paper ID \FGPaperID \\}
\pagestyle{plain}
\fi
\maketitle

\thispagestyle{fancy}

%%%%%%%%%%%%%%%%%%%%%%%%%%%%%%%%%%%%%%%%%%%%%%%%%%%%%%%%%%%%%%%%%%%%%%%%%%%%%%%%
\begin{abstract}

Generative 3D face models featuring disentangled controlling factors hold immense potential for diverse applications in computer vision and computer graphics. However, previous 3D face modeling methods face a challenge as they demand specific labels to effectively disentangle these factors. This becomes particularly problematic when integrating multiple 3D face datasets to improve the generalization of the model. Addressing this issue, this paper introduces a Weakly-Supervised Disentanglement Framework, denoted as \ShockingName{}, to facilitate the training of controllable 3D face models without an overly stringent labeling requirement. Adhering to the paradigm of Variational Autoencoders (VAEs), the proposed model achieves disentanglement of identity and expression controlling factors through a two-branch encoder equipped with dedicated identity-consistency prior. It then faithfully re-entangles these factors via a tensor-based combination mechanism. Notably, the introduction of the Neutral Bank allows precise acquisition of subject-specific information using only identity labels, thereby averting degeneration due to insufficient supervision. Additionally, the framework incorporates a label-free second-order loss function for the expression factor to regulate deformation space and eliminate extraneous information, resulting in enhanced disentanglement. Extensive experiments have been conducted to substantiate the superior performance of \ShockingName{}.
Our code is available at \color{gray}{\href{https://github.com/liguohao96/WSDF}{github.com/liguohao96/WSDF}}.

\end{abstract}

%%%%%%%%%%%%%%%%%%%%%%%%%%%%%%%%%%%%%%%%%%%%%%%%%%%%%%%%%%%%%%%%%%%%%%%%%%%%%%%%
\section{INTRODUCTION}
3D face modeling is a persistent topic due to its numerous applications in discriminative and generative tasks within 2D and 3D domains, such as
2D face recognition \cite{Liu2020:Joint,Yang2022:BareSkinNet},
facial attribute transfer \cite{Thies2018:Face2Face},
3D face reconstruction \cite{Deng2019:Accurate}
and virtual avatars \cite{hong2021headnerf,sun2023next3d}.
The versatility of 3D face modeling stems from its controllable and generative capabilities, enabling the representation and generation of 3D faces with disentangled, interpretable, and editable factors.

Linear 3D Morphable Models (3DMMs) \cite{3DMM:99,3DMM:03,Amberg2008:ExpressionInvariant,BFM2017} achieve controllable generation of facial geometry by separately modeling identity and expression deformation through Principal Component Analysis (PCA) applied to 3D facial scans annotated with both identity and expression labels. To enhance performance, dense expression labels are utilized in constructing multi-linear models \cite{bilinear,bilinear:siggraph,FaceWarehouse}. Unfortunately, the limited representation ability of linear models and the scarcity of expression-annotated 3D facial scans hinder the generalizability of these methods. The advent of deep learning has spurred the exploration of non-linear models, employing explicit \cite{CompositionalVAEs,CoMA,Neural3DMM,Liu2019:3DFace,Danecek2022:EMOCA} or implicit modeling \cite{Yenamandra2021:i3DMM,Zheng2022:ImFace,Zhang2023:NPF}, leveraging training data from both 2D images \cite{Tran2018:Nonlinear,Tran2019:Towards,Tran2021:OnLearning} and 3D scans \cite{Liu2019:3DFace,Bahri2021:SMF} for improved generalizability. Nevertheless, the persistent need for expression labels \cite{Tewari2019:FML,R2021:Learning,Liu2019:3DFace,Zheng2022:ImFace,sun2023next3d} to achieve identity and expression disentanglement remains an unresolved challenge. Since 3D face datasets are often collected and labeled for diverse purposes, and joint expression labels are typically unavailable, existing methods are confined to a limited number of 3D facial datasets with dedicated expression labels. This limitation becomes a significant obstacle when attempting to combine multiple datasets to enhance diversity, ultimately compromising the overall generalizability.

Recent studies have demonstrated the remarkable capability of generative models in disentangling latent spaces \cite{Higgins2017:betaVAE,chen2018:TCVAE,kwon2021:diagonal,yue2022anifacegan,sun2023next3d} . Furthermore, insights from \cite{Locatello2019:Challenging} underscore the importance of introducing inductive biases on data or models for effective representation disentanglement. Notably, these studies propose an alternative training scheme for 3DMMs, suggesting that the need for explicit expression labels can be circumvented. This is achieved by treating facial expression as a complementary, identity-irrelevant factor, leveraging the abundance of identity information in almost all existing 3D face datasets. To our knowledge, Gu \textit{et al.} are pioneers in exploring such inductive bias to separate identity and expression latent spaces \cite{gu2023adversarial}. However, it is worth noting that their use of adversarial training to implicitly learn identity information is not foolproof in preventing local minima arising from the ambiguity of weak supervision. Both whole and partial identity-relevant information may satisfy identity-consistency, as highlighted in \cite{Shu2020:Weakly,gu2023adversarial}.

This paper focuses on advancing 3D facial shape modeling and introduces an innovative 3D Morphable Model (3DMM) that effectively disentangles identity and expression representations using only weak supervision provided by identity labels. 
% 3D representation
Our approach utilizes a topology-consistent mesh to represent 3D geometries, chosen for its efficiency in generation, editing, and rendering tasks. The model adopts a VAE-based paradigm, employing two encoders to learn identity and expression latent spaces and re-coupling them during decoding to generate 3D faces. 
% neutral bank & neu loss
A key feature of our model is its handling of the ambiguity in identity representations when relying solely on identity-consistency. To address this, we introduce dedicated constraints to achieve disentanglement and impose an identity-consistency prior. Specifically, we introduce the Neutral Bank module, which obtains pseudo-neutral scans for each subject by aggregating multiple samples. This module enhances the consistency of the identity branch by retaining identity-related information through the maximization of mutual information.
% jacobian loss
For the expression branch, we employ a second-order loss with regularized morphing energy to induce encoding variations of expression in diverse directions within the latent space. This strategy effectively removes extraneous information from the expression space. Additionally, we propose a non-linear tensor-based fusion mechanism to combine these disentangled latent spaces, accurately modeling subject-specified expressions.

In summary, this study makes several key contributions:
\begin{itemize}

\item \textbf{\NBankName{} Module:} We introduce a \NBankName{} module, complemented by a dedicated loss function. This module plays a crucial role in preventing the degeneration of identity-consistency, thereby facilitating the disentanglement of the identity factor.

\item \textbf{Label-Free Second-Order Loss: }
We propose a label-free second-order loss, designed to enhance disentanglement by eliminating nuisance information within the expression space. This is achieved through the regularization of deformation, resulting in more effective disentanglement.

\item \textbf{A novel 3DMM:}
We present a novel 3DMM, namely \ShockingName{}, which is learned from combined datasets. Experimental validation demonstrates its significantly improved generalizability, marking it as an advancement in the field of 3D facial shape modeling.
\end{itemize}

%%%%%%%%%%%%%%%%%%%%%%%%%%%%%%%%%%%%%%%%%%%%%%%%%%%%%%%%%%%%%%%%%%%%%%%%%%%%%%%%
\section{RELATED WORK}
% \subsection{3D Face Modeling}
\textbf{3D Face Modeling.}
\cite{3DMM:99,3DMM:03} 
propose generative 3DMM that models 3D human faces with disentangled factors facilitating controllability.
The vanilla 3DMM separately models geometry and appearance of human face with two sets of linear bases. % learned by performing PCA on 200 neutral expression facial scans and their corresponding textures.
Subsequent 3DMMs \cite{Amberg2008:ExpressionInvariant,BFM2009,BFM2017,flame,Albedo3DMM} are proposed to improve the controllability of generative modeling by disentangling more diverse factors with dedicated supervisions.
While controllability is essential, another line of interest is generalizability, which involves both realistic reconstruction and diverse generation.
To improve this ability, a large amount of training data are typically required.
A large scale face model (LSFM) \cite{LSFM} is build from non-public scans of 9,663 subjects within neutral expression.
The most commonly used FLAME model, build by \cite{flame}, uses a total 33,000 3D scans from 4D videos with the aid of 3D-artist.
Different from the studies above, \cite{Ploumpis2019:Combining} circumvent the scarcity of 3D facial scans by combining multiple 3D face models.
However, the generalizability is restricted by model linearity and inability to automatically learn from large-scale unlabeled scans.

% \textbf{Non-Linear 3D Face Modeling}
More recently, the development of neural networks has facilitated the creation of nonlinear models for 3D facial geometry.
By adopting differentiable rendering,
3DMMs can be learned from in-the-wild 2D images \cite{Tran2018:Nonlinear,Tran2019:Towards,Tran2021:OnLearning}
through solving inverse render problems during training.
The models can also be trained in an end-to-end manner on 3D scans using an encoder-decoder framework \cite{CompositionalVAEs,Liu2019:3DFace,Bahri2021:SMF,Danecek2022:EMOCA}.
Furthermore, several methods based on Implicit Neutral Representations (INR) 
have been proposed \cite{Yenamandra2021:i3DMM,Zheng2022:ImFace,Galanakis2023:3DMMRF,Zhang2023:NPF}.
% employing Signed Distance Fields (SDFs) \cite{SDF} or Neural Radiance Fields (NeRFs) \cite{NeRF} to learn facial geometry and appearance properties.
Despite achieving realistic reconstruction with powerful representations,
these methods still require expression labels to disentangle geometry into identity and expression for controllability.

% \subsection{Weakly-supervised Disentanglement}

% \textbf{Weakly-supervised Disentanglement for Generative Model} 
% As \cite{Locatello2019:Challenging} demonstrate that learning disentanglement
% without inductive bias of data or model is impossible,
% weakly-supervised disentanglement has drawn a lot attention.
% $\beta$-VAE, FactorVAE, GAN inversion.

\textbf{3D Face Disentanglement.}
Recent studies on 3D face disentanglement focused on semantically disentangling 3D face into identity and expression factors.
Several approaches \cite{Lin2022:SDVAE,olivier2023facetunegan}
relying on dense annotations for decoupled factors have been proposed.
Since labels are dataset-specified, however, 
these methods are not applicable to datasets with only sparse expression labels.
Approaches \cite{Liu2019:3DFace,zhang2020learning,Bahri2021:SMF,Wang2022:FaceVerse,
kacem2022disentangled,Zheng2022:ImFace,sun2022information}
relax the demand for expression annotations and use only \emph{neutral} labels as an inductive bias.
They assumes that variations of neutral scans solely attribute to identity-relevant factors.
However, 
the existence of neutral scans is still correlated with datasets,
and some datasets lack such \emph{neutral} label, \textit{e.g.}  CoMA\cite{CoMA}.
\cite{gu2023adversarial} employ a discriminator to learn the inductive bias 
from identity-consistency and bypass the need for neutral scans,
which is similar to our assumption.
However, their method suffers from degenerated solutions, as shown in their experiments.
Moreover, the fact that expressions are identity-specified
is neglected in their additive combination of identity and expression.
We prevent degeneration with information-preserving mechanism
and adopt tensor-based combination to model subtle differences of the same expression category perform by different individuals.
\section{METHOD}
\begin{figure*}[t]
    \centering
    \includegraphics[width=0.9\linewidth]{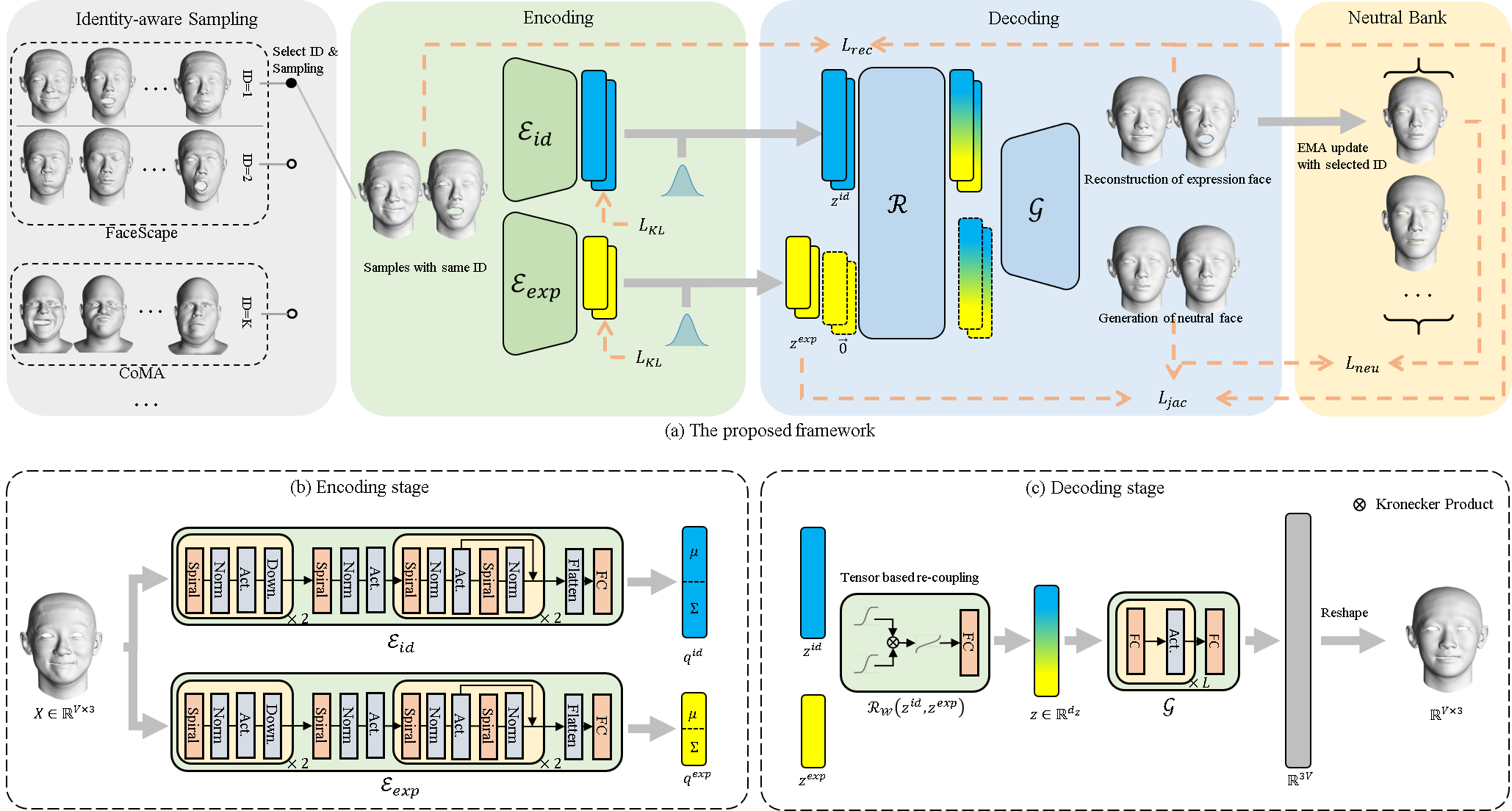}
    \caption{Method overview.
    (a) Identity-aware sampling constructs a group of training samples belonging to the same ID.
    Two encoders, $\script{E}_{id}$ and $\script{E}_{exp}$ are built to disentangle identity and expression representations.
    The disentangled latent codes, $\codei$ and $\codee$, are re-coupled by fusion module $\script{R}$ and fed into the generator $\script{G}$ for decoding.
    Simultaneously, a neutral bank is constructed to obtain pseudo-neutral scans for each subject, imposing disentanglement through the $\script{L}_{neu}$ loss.
    (b) The encoders  $\script{E}_{id}$ and $\script{E}_{id}$ utilize the Spiral++ architecture with the same setup, aiming to factor out the inductive bias of model design. 
    (c) The decoder involves a tensor-based Recoupler $\script{R}$ with non-linearity and an MLP-based $\script{G}$ for mapping from decoupled latent codes  $\codei$ and $\codee$ to 3D faces.
    }
    \label{fig:framework}
\end{figure*}

This paper presents \ShockingName{}, a Weakly-supervised Disentanglement Network designed to learn a generative 3D face model controlled by identity and expression factors independently. The approach is based on the inductive bias of identity-consistency, which relaxes the requirements for expression annotations. In the proposed framework, to eliminate the inductive bias of model design, a two-branch encoder with an identical architecture is designed to learn disentangled latent spaces. The combination of identity and expression is achieved through a non-linear tensor-based mechanism, providing an accurate representation of subject-specified expressions. To disentangle identity, the introduction of a \NBankName{} module is pivotal, generating pseudo-neutral scans using identity labels. This module serves as supervision to enforce identity-consistency and information preservation simultaneously. To eliminate nuisance information in the expression space, a second-order loss is proposed. This loss encourages the encoding of expression variations in diverse directions and regulates the energy of these variations, contributing to improved disentanglement. Overall, \ShockingName{} presents an innovative approach to weakly-supervised disentanglement in 3D face modeling. In the following sections, we provide a detailed description of our proposed framework and loss functions.

\subsection{Framework}

As shown in Fig.~\ref{fig:framework}(a), the proposed framework is built upon mesh-based VAEs.
Two encoders, $\script{E}_{id}$ and $\script{E}_{exp}$,
are adopted to learn the decoupled latent spaces.
The latent spaces are then re-coupled by tensor-based \MixerName{} $\script{R}$
and fed into a generative network $\script{G}$ for reconstruction and generation.
In particular, a group of 3D scans $\{X_i\}$ belonging to the same subject is sampled 
from scans gathered according to IDs before training.
This encoding stage could be formulated as:
$\distg{i} = \script{E}_{id}(X_i),\diste{i} = \script{E}_{exp}(X_i)$,
where 
$\distg{}$ and $\diste{}$ denote the distributions produced by $\script{E}_{id}$ and $\script{E}_{exp}$, respectively.
In the context of VAE, re-parameterization technique is utilized to
draw identity latent code $\codei$ and expression latent code $\codee$ for each scan from its corresponding $\disti{}$ and $\diste{}$, respectively.
As shown in Fig.~\ref{fig:framework}(b), $\script{E}_{id}$ and $\script{E}_{exp}$ leverage Spiral++ architecture~\cite{Gong2019:Spiral} with Instance Normalization~\cite{ulyanov2017instance} and ELU activation~\cite{elu}.

\newcommand{\mugo}[1][]{\mu^{\scriptscriptstyle  \nameg}_#1}
\newcommand{\vargo}[1][]{\Sigma^{\scriptscriptstyle  \nameg}_#1}              %\newcommand{\ld}{i\in \gn}      
\newcommand{\muid}[1][]{\mu^{\scriptscriptstyle \namei}_#1}              %\newcommand{\ld}{i\in \gn}      
\newcommand{\varid}[1][]{\Sigma^{\scriptscriptstyle \namei}_#1}              %\newcommand{\ld}{i\in \gn}      
\newcommand{\muexp} {{\mu^{    \scriptscriptstyle exp}   }}              %\newcommand{\ld}{i\in \gn}      
\newcommand{\varexp}{{\Sigma^{ \scriptscriptstyle exp}}}              %\newcommand{\ld}{i\in \gn}      

\newcommand{\gn}{\namei}
\newcommand{\gs}{\vert{\{\mu_{\nameg}\}}\vert}
\newcommand{\lu}{\vert G \vert}    %\newcommand{\lu}{}               
\newcommand{\ld}{i=1}              %\newcommand{\ld}{i\in \gn}      

\newcommand*{\tmod}[3]{#2\times_{#1}#3}
\newcommand*{\modn}[1]{\times_{#1}}
\newcommand*{\matU}[2][]{U_{#1}^{(#2)}}
\newcommand*{\asmatrix}[1]{\double{M}_{#1}}
\newcommand*{\tonormal}[1]{\double{N}_{#1}}
\newcommand*{\touniform}[1]{\double{U}_{#1}}
\newcommand{\tdoneo}{r}
\newcommand{\tdonei}{i}
\newcommand{\tdonen}{I}
\newcommand{\tdtwoo}{s}
\newcommand{\tdtwoi}{j}
\newcommand{\tdtwon}{J}

We design a two-stage decoding pipeline, as shown in Fig.~\ref{fig:framework}(c).
In the first stage, the latent codes are combined together by tensor-based multiplications with non-linear functions through \MixerName{} $\script{R}$.
In the second stage, a generator network $\script{G}$ is adopted to recover variations of 3D faces.
This decoding stage could be formulated as:
\begin{equation}
    \label{eq:decode}
    \begin{split}
    \xrec{i} = \script{G}\circ\script{R}(\codei_i, \codee_i),\
    \xneu{i} = \script{G}\circ\script{R}(\codei_i, \vec{0}),
    \end{split}
\end{equation}
where $\xrec{i}$ and $\xneu{i}$ denote the reconstruction and neutralization results of input scan $X_i$, respectively.
To maintain compatibility with previous 3DMMs, a zero vector $\veczero$ is 
utilized to generated neutralized expression (de-expression) faces.

\textbf{Re-coupling of Factors.}
To efficiently represent subject-specified expressions,
we leverage a tensor-based bilinear fusion
% similar to multi-linear 3DMMs \cite{Egger2020:3DMMPPF},
with dedicated nonlinear functions to ensure the re-coupled latent code $z$ follows normal distribution.
Without loss of generality, the tensor-based bilinear fusion could be described as
$\script{W}\modn{2}\matU{2}\modn{3}\matU{3}$
where 
$\script{W}\in\double{R}^{k\times\tdonei \times \tdtwoi}$ is the weight \emph{tensor},
$\matU{2}\in\double{R}^{\tdoneo \times \tdonei}$ and 
$\matU{3}\in\double{R}^{\tdtwoo \times \tdtwoi}$ are parameter \emph{matrices}
and $\times_{n}$ denotes the \emph{mode-n} product.
Despite its higher-dimensional representation, this bilinear model is equivalent to \emph{matrix} multiplication as shown below:
\begin{displaymath}
    \begin{split}
    (\script{W} \modn{2} \matU{2} \modn{3} \matU{3})_{k\tdoneo \tdtwoo} &=
        \sum_{\tdtwoi =1}^{\tdtwon}{(\script{W} \modn{2} \matU{2})_{k\tdoneo \tdtwoi}} \matU[\tdtwoo \tdtwoi]{3}\\
        &=\sum_{\tdtwoi =1}^{\tdtwon}{\sum_{\tdonei =1}^{\tdonen}{\script{W}_{kij}}\matU[\tdoneo \tdonei]{2}}\matU[\tdtwoo \tdtwoi]{3}\\
        &=\asmatrix{\script{W}} (\matU{2} \otimes \matU{3})^T,\\
    \end{split}
\end{displaymath}
where $\asmatrix{\script{W}}\in\double{R}^{k\times ij}$ denotes the reshaped \emph{matrix} of \emph{tensor} $\script{W}$.
When it comes to our \MixerName{}, $\matU{2}$ and $\matU{3}$ are replaced by the disentangled \emph{vectors} $\codei_i$ and $\codee_i$, respectively.
% Since $\codei\sim\distnormal(\mu_{\namei},\Sigma_{\namei})$, $\codee\sim\distnormal(\mu_{exp},\Sigma_{exp})$ and the product of two gaussian random variables follows a complicated distribution,
Since elements in $\codei$ and $\codee$ follow a normal distribution $\distnormal(0, 1)$, and the product of two gaussian random variables follows a complicated distribution,
the multiplied result will not preserve normal distributions.
To address this issue, we adopt inverse transform sampling technique: %in \MixerName{}:
% $\codei$ and $\codee$ are transformed into $\touniform{\codei}\sim\distuniform(0,1)$ and $\touniform{\codee}\sim\distuniform(0,1)$
$\codei$ and $\codee$ are transformed into $\touniform{\codei},\touniform{\codee}\sim\distuniform(0,1)$
first by non-linear function $\touniform{}$, then the multiplied result $\touniform{\codei} \otimes \touniform{\codee}$ is transformed into $\tonormal{\touniform{\codei} \otimes \touniform{\codee}}\sim\distnormal(0,1)$
by non-linear function $\tonormal{}$.
Finally, these normal distribution variables are linear combined by $\asmatrix{\script{W}}$. %reshaped weight tensor $\asmatrix{\script{W}}$.
Although the elements of $\tonormal{\touniform{\codei} \otimes \touniform{\codee}}$ are dependent,
we treat them as independent ones and use a normalized $\asmatrix{\script{W}}$ to ensure the variance of $z$ is 1.
Formally, the proposed \MixerName{} is defined as:
\begin{equation}
    \script{R}_{\script{W}}(\codei,\codee)=\asmatrix{\script{W}} \tonormal{{\touniform{\codei}} {\otimes} {\touniform{\codee}}}^T.
\end{equation}
Thanks to the preserved normal distribution, any pre-trained generative model follows this prior could be used as $\script{G}$.

\textbf{\NBankName{}.}
Since identity-aware sampling guarantees all samples in $\{X_i\}$ share the same identity, the distribution $\{\disti{i}\}$ should be the same according to identity-consistency.
Thus, the neutralized 3D face $\{\xneu{i}\}$ should roughly be the same.
However, enforcing similarity to fulfill identity-consistency alone can be ambiguous regarding the amount of retained identity-relevant information,
which leads to a myriad of local minima.
As a consequence,
the unconstrained $\codee$ learns to preserve the majority of identity-relevant information,
while $\codei$ retains less identity-relevant information.
This phenomenon is similar to what was observed in \cite{gu2023adversarial} and stated in \cite{pang2023dpe}.
This issue could be addressed by maximizing mutual information of $\codei$ and the underlying identity factors. 
%,with the inspiration of the information bottleneck.
% Specifically, $\codei$ requires dedicated inductive biases in the context of weak supervision.

We consider scans with neutral expression retrain only identity-related information and propose to learn such scans with dedicated inductive biases in the context of weak supervision.
Assuming that expression morphing is additive to the neutral face \cite{Zhang2021:Learning,Egger2020:3DMMPPF}, 
a neutral scan $s$ from $n$ scans of same subject $\{X_i\}$ could be learned through solving
an under-determined linear system with least-square constraint:
\newcommand{\matW}{\begin{bmatrix}
    s \\ \delta_1 \\ \vdots \\ \delta_n
\end{bmatrix}}
\newcommand{\matWt}{\begin{bmatrix}
    s & \delta_1 & \cdots & \delta_n
\end{bmatrix}}
\newcommand{\matA}{\left [ \begin{array}{c|c}
    {1}_{n\times 1} & {I}_{n\times n}
\end{array} \right ] }
\newcommand{\matB}{\begin{bmatrix}
    {X}_1 \\ \vdots \\ {X}_n
\end{bmatrix}}
\begin{displaymath}
    \begin{split}
        \argmin{ \sum_{i}{\norm{\delta_i}} } & \matA \matW = \matB
    \end{split},
\end{displaymath}
% where $s$ is the neutral face, $\delta_i$ denotes expression-related offsets from $s$ to $i$-th scan $\setof{X}_i$.
where $s$ is the neutral face, $\delta_i$ denotes expression-related offsets from $s$ to the $i$-th scan ${X}_i$.
%It is important to note that $s$, $\delta_i$ and $\setelement[i]{X}$ are matrices and abused for simplicity without the loss of generality.
These sparse least-square problem could be solved in a closed form as:
$s = \frac{1}{n}\sum_{i}^{n}\setelement{X}$,
$\delta_i = \setelement{X} - s$,
% \begin{displaymath}
%     \begin{split}
%         s &= \frac{1}{n}\sum_{i}^{n}\{X\}_i, \\
%         \delta_i &= \{X\}_i - s,
%     \end{split}
% \end{displaymath}
where $\delta_i$ could be effected by both identity and expression.
This suggests that an averaging of observed scans can approximately serve as the pseudo neutral scan.
In the proposed method, we construct a \NBankName{} module which stores the pseudo neutral scan for each subject and leverages the Exponential Moving Average (EMA) to learn on the fly using the reconstructed faces $\{\xrec{i}\}$ belonging to the same subject.
Given the selected subject ID, $K$, during identity-aware sampling and current step $t$,
the updating of \NBankName{} module can be formulated as:
\begin{equation}
\script{B}_{K}^{(t+1)}=\beta\script{B}_{K}^{(t)}+(1-\beta)\overline{\xrec{}},
\end{equation}
where $\beta$ is the hyper-parameter of EMA and $\overline{\xrec{}}$ denotes the average of $\{\xrec{i}\}$.
The pseudo neutral scans for subject $K$, $\script{B}_{K}$, are utilized to maximize the identity-relevant information preserved in $\codei$
through the neutralized faces $\setof{\xneu{i}}$.
Since all samples in $\setof{\xneu{i}}$ are generated using constant expression code, as in Eq.~\ref{eq:decode},
the subject-specific information could only be encoded by learnable $\codei$.
In this way, with only identity label, $\codei{}$ learns to preserve identity-related information without ambiguity and achieves identity-consistency implicitly with subject specified $\script{B}_{K}$.
It is worth noting that this approach introduces the inductive bias of face deformation (additive) and energy-minimizing (least-square constraint).
Compared to using averaged scans as $\script{B}$, learning from EMA updated $\script{B}$ follows a progressive learning principle and \NBankName{} module may benefit from online expression augmentation.
Besides, thanks to the agnostic of input topology, the proposed \NBankName{} module could be extended to deal with heterogenous scans.
% In this way, the degenerated solution can be largely prevented and consistency across $\{\codei_i\}$ is imposed implicitly. It is worth noting that this solution introduces the inductive bias of face deformation (additive) and energy-minimizing (least-square constraint).

% \newcommand*{\DR}{\citet{jiang2019disentangled}}
% \newcommand*{\DIMeshEnc}{\citet{zhang2020learning}}
% \newcommand*{\FTG}{FaceTuneGAN \cite{olivier2023facetunegan}}
% \newcommand*{\IBVAE}{IB-VAE \cite{sun2022information}}
% \newcommand*{\AFD}{FED \cite{gu2023adversarial}}
\newcommand*{\DR}{DRL}
\newcommand*{\DIMeshEnc}{DI-MeshEnc}
\newcommand*{\FTG}{FaceTuneGAN}
\newcommand*{\IBVAE}{IB-VAE}
\newcommand*{\AFD}{FED}

\definecolor{Gray}{gray}{0.9}
\definecolor{LightGray}{gray}{0.6}
\newcommand*{\cg}{\color{LightGray}}
\gdef\ul#1{\underline{#1}}

\begin{table*}[h]
  % \resizebox*{0.9\textwidth}{!}{
  \centering
  \caption{Quantitative comparison with other methods on the FaceScape and CoMA datasets.}
  \label{tab:dataset_separate}
    \begin{tabular}{ccccccccc}
        \toprule
        \multirow{2}{*}{Dataset} & \multirow{2}{*}{Method} & \multirow{2}{*}{Exp. Label} & \multicolumn{2}{c}{$E_{avd}$} & \multicolumn{2}{c}{$E_{id}$} & \multicolumn{2}{c}{$E_{neu}$} \\
                                                  &        &                                &      mean\lowb & median\lowb  &     mean\lowb & median\lowb  &     mean\lowb & median\lowb   \\
        \midrule
        \multirow{5}{*}{FaceScape} 
                                   & \FTG{}                  & Dense   & \cg\rpfloat[2]{0.81}                 
                                                                       & \cg\rpfloat[2]{0.76}
                                                                       & -                                    
                                                                       & -                     
                                                                       & \cg\rpfloat[2]{1.42}                 
                                                                       & \cg\rpfloat[2]{1.26}   \\  % table 8 & table 4
                                   & \AFD{}                  & Sparse  &    \rpfloat[3]{1.370}                   
                                                                       &    \rpfloat[3]{1.307}
                                                                       &    \rpfloat[3]{0.569}                   
                                                                       &    \rpfloat[3]{0.551}
                                                                       &    \rpfloat[3]{1.927}                   
                                                                       &    \rpfloat[3]{1.815} \\  % table 3
                                   \cline{2-9}
                                   & \AFD{} w/o exp. label   & -       &    \rpfloat[3]{1.157}                   
                                                                       &    \rpfloat[3]{1.109}   
                                                                       &    \rpfloat[3]{0.765}                   
                                                                       &    \rpfloat[3]{0.758}
                                                                       &    \rpfloat[3]{3.112}                   
                                                                       &    \rpfloat[3]{2.957} \\  % table 3
                                   & \bf Ours                & -       & \bf{\rpfloat[3]{1.110}}  
                                                                       & \bf{\rpfloat[3]{0.998}}
                                                                       & \bf{\rpfloat[3]{0.366}} 
                                                                       & \bf{\rpfloat[3]{0.353}}
                                                                       & \bf{\rpfloat[3]{1.590}}       
                                                                       & \bf{\rpfloat[3]{1.485}} \\ % C3 F-L1-nb1-jac10 FaceScape:100% @1407*100
        \midrule
        \multirow{7}{*}{CoMA}   
                                   & \FTG{}                  & Dense   & \cg\rpfloat[2]{0.83 }                
                                                                       & \cg\rpfloat[2]{0.77 }
                                                                       & \cg\rpfloat[3]{0.018}                
                                                                       & \cg\rpfloat[3]{0.018}
                                                                       & \cg\rpfloat[2]{0.98}                 
                                                                       & \cg\rpfloat[2]{0.81}  \\  % table 1 & table 3 & table 4
                                   & \DR{}                   & Dense   & \rpfloat[3]{1.413}                   
                                                                       & \rpfloat[3]{1.017}
                                                                       & \rpfloat[3]{0.064}                   
                                                                       & \rpfloat[3]{0.062}    
                                                                       &  -                                   
                                                                       & - \\
                                   & \DIMeshEnc{}            & Dense   &    {\rpfloat[3]{0.665}}                   
                                                                       &    {\rpfloat[3]{0.434}}
                                                                       &    {\rpfloat[3]{0.019}}                   
                                                                       &    {\rpfloat[3]{0.020}}    
                                                                       &  -                                   
                                                                       & - \\
                                   & \IBVAE{}                & Sparse  &    {\rpfloat[3]{0.663}}                   
                                                                       &    {\rpfloat[3]{0.643}}
                                                                       &    \rpfloat[3]{0.006}                
                                                                       &    \rpfloat[3]{0.005}
                                                                       & \cg\rpfloat[3]{0.051}                
                                                                       & \cg\rpfloat[3]{0.048} \\ % table 1 & table 2
                                   & \AFD{}                  & Sparse  &    {\rpfloat[3]{0.651}}                   
                                                                       &    {\rpfloat[3]{0.625}}
                                                                       &    {\rpfloat[3]{0.014}}                   
                                                                       &    {\rpfloat[3]{0.013}}
                                                                       & \cg{\rpfloat[3]{0.065}}                   
                                                                       & \cg{\rpfloat[3]{0.063}} \\ % table 1
                                   \cline{2-9}
                                   & \AFD{} w/o exp. label   & -       &    {\rpfloat[3]{0.783}}                   
                                                                       &    {\rpfloat[3]{0.772}}
                                                                       &    {\rpfloat[3]{0.176}}                   
                                                                       &    {\rpfloat[3]{0.180}}
                                                                       &    {\rpfloat[3]{1.528}}                   
                                                                       &    {\rpfloat[3]{1.268}} \\ % table 1 & table 4
                                   & \bf Ours                & -       & \bf{\rpfloat[3]{0.480}}      
                                                                       & \bf{\rpfloat[3]{0.409}} 
                                                                       & \bf{\rpfloat[3]{0.005}} 
                                                                       & \bf{\rpfloat[3]{0.002}}  
                                                                       & \bf{\rpfloat[3]{0.651}}  
                                                                       & \bf{\rpfloat[3]{0.697}}    \\ % C3 C-L3-neu10-jac10-mi10-rerun1 CoMA:100% @2187*100
        \bottomrule
    \end{tabular}
  % }
\end{table*}

\subsection{Loss Functions}
The proposed generative framework learns to model 3D faces with disentangled factors in a fashion of VAE.
Besides maximizing the Evidence Lower Bound (ELBO) of VAE, auxiliary losses are proposed to disentangle latent spaces through information bottleneck-based mechanism and label-free bijectivity mapping constraint.

\textbf{Reconstruction Loss}.
With the assumption that all facial scans are registered in the same topology,
the likelihood term in ELBO could be defined
as reconstruction loss between the input scan $X$ and the decoded face $\xrec{}$:
\begin{equation}
    \begin{split}
        \script{L}_{rec} &= \norm{X-\xrec{}}^2.
    \end{split}
\end{equation}

\textbf{KL Loss}.
As for Kullback-Leibler (KL) divergence term in ELBO, we impose KL divergence
on two decoupled distributions $\disti{}$ and $\diste{}$ as:
\begin{equation}
    \begin{split}
        \script{L}_{KL} &= KL( \disti{}\ \vert\vert\ \distnormal(0, I)) + KL( \diste{}\ \vert\vert\ \distnormal(0, I)).
    \end{split}
\end{equation}

\textbf{Neutralization Loss}.
To prevent degeneration of the identity latent space and enforce identity-consistency,
the mutual-information between $\codei$ and the underlying identity information is maximized
by using the generated neutralized faces $\setof{\xneu{}}$.
We utilize pseudo neutral scans, $\script{B}_{K}$,
of $K$-th subject from \NBankName{} as a representation of underlying identity information.
Similar to $\xrec{}$, likelihood is maximized by a reconstruction loss, defined as:
\begin{equation}
    \begin{split}
        \script{L}_{neu} &= \alpha_{K}\norm{\script{B}_{K}-\xneu{}}^2,\ 
        \alpha_K=1-\exp(1-\vert {K} \vert)
    \end{split}
\end{equation}
where $\vert {K} \vert$ represents the number of samples for subject $K$.
The factor $\alpha_{K}$ can be interpreted as the confidence score of EMA updated $\script{B}_{K}$,
since the EMA updated $\script{B}_{K}$ can hardly guarantee to preserve only identity-related information when few variety of expressions of the same subject are demonstrated during training.
Identity-consistency across $\{\disti{i}\}$ is implicitly learned through reconstruction the same target, 
which factors expression out of identity space.

\newcommand*{\jaco}[1]{J_{#1}}
\newcommand*{\normxtx}[1]{{#1}^T{#1}}
\newcommand*{\var}{y}

\textbf{\JLossName{}.}
To achieve better disentangled expression representations without label,
we enforce the mapping from the \emph{norm} of expression latent code $\codee$
to the intensity of expression is bijective and leverage energy-minimizing prior.
Thus, diverse expressions under the same intensity should be mapped to different directions in $\codee$;
otherwise, the decoder is unable to distinguish these expressions.
We model the intensity of expression for $\xrec{}$ as the \emph{norm} of $\xrec{}-\xneu{}$.
Therefore, this bijective mapping could be described
by a non-negative monotonically increasing scalar function $\sigma$:
        % &\exists \sigma:\double{R}_{+}\rightarrow\double{R}_{+},\ \sigma^{'}>0,\\
$\sigma^{'}>0$,
$s.t.\ \normxtx{(f(\var)-f(\veczero))} =\sigma(\normxtx{\var})$,
% \begin{displaymath}
%     \begin{split}
%         &\exists \sigma,\ \sigma^{'}>0,\\
%         s.t.\ &\normxtx{(f(\var)-f(\veczero))} =\sigma(\normxtx{\var}),
%     \end{split}
% \end{displaymath}
where $\var$ indicates $\codee$ and $f(\var)=\script{G}\circ\script{R}(\codei,\var)$.
The subject-specified $\sigma$ and identity code $\codei$ is omitted for clarity.
Since it is hard to explicitly formulate $\sigma$, we compute its equivalent form:
%compute $\frac{d}{d\var{}}$ on both side and 
$(f(\var{}) - f(\veczero))^T\jaco{f}(\var{})\var{} = \sigma{'}(y^Ty)y^Ty$,
% $(f(\var{}) - f(\veczero))^T\jaco{f}(\var{})\var{} \geq 0$,
% Since it is hard to explicitly formulate $\sigma$, we transform this equation by:
% \newcommand*{\jaco}[1]{J_{#1}}
% \newcommand*{\normxtx}[1]{{#1}^T{#1}}
% \begin{displaymath}
%     \begin{split}
%         \frac{d}{d\var{}} on\ both\ side \Rightarrow &(f(\var) - f(\veczero))^T\jaco{f}(\var)     = \sigma^{'}(\var{}^T\var{})\var{}^T, \\
%         multiply\ \var{}\ on\ both\ side \Rightarrow &(f(\var) - f(\veczero))^T\jaco{f}(\var)\var = \sigma^{'}(\var{}^T\var{})\var{}^T\var{}, \\
%   \Rightarrow &(f(\var{}) - f(\veczero))^T\jaco{f}(\var{})\var{} \geq 0,
%     \end{split}
% \end{displaymath}
where $\jaco{f}(\var)$ stands for jacobian matrix.
This formula suggests that we could 
further improve disentangled representation learning with energy-minimizing prior by enforcing $0\leq (f(\var) - f(\veczero))^T\jaco{f}(\var)\var \leq \sigma^{'}_{max}y^Ty$.
The lower bound enforces a bijectivity mapping which necessitates the incorporation of diverse directions in encoding expressions, 
while the overall deformation is regularized by the locally defined upper bound.
Consequently, the expression space is restricted to exclusively model expression deformation.
In this case, instead of finding subject specified $\sigma^{'}_{max}$, we constrain it with a hyper parameter $\gamma > 0$.
We implement such constraint as:
\begin{equation}
    \renewcommand*{\jaco}[1]{J_{#1}}
    \renewcommand*{\normxtx}[1]{{#1}^T{#1}}
    \begin{split}
        \script{L}_{jac} = max(&0, -(\xrec{}-\xneu{})^T\jaco{\codee}\codee, \\
        &(\xrec{}-\xneu{})^T\jaco{\codee}\codee - \gamma {{\codee}^T}\codee),
    \end{split}
\end{equation}
where $\jaco{\codee}$ is the jacobian matrix with respect to $\codee$, given input $\codei$ and $\codee$.
Thanks to the MLP generator network $\script{G}$, the jacobian matrix could be computed efficiently.

\textbf{Identity Compactness Loss.}
We also use $\script{L}_{mi}$ proposed in \cite{sun2022information} to enforce compactness of identity branch.
Our overall loss function is:
\begin{equation}
    \begin{split}
        \script{L} &= \script{L}_{rec} + 
                      \script{L}_{KL}  + 
                      \lambda_{neu} \script{L}_{neu} +
                      \lambda_{jac} \script{L}_{jac} +
                      \lambda_{mi}  \script{L}_{mi}.
    \end{split}
\end{equation}
where $\lambda_{neu}$, $\lambda_{jac}$ and $\lambda_{mi}$ are hyper-parameters that control and balance different losses.

%%%%%%%%%%%%%%%%%%%%%%%%%%%%%%%%%%%%%%%%%%%%%%%%%%%%%%%%%%%%%%%%%%%%%%%%%%%%%%%%
\section{EXPERIMENT}
\subsection{Datasets}
We evaluate the effectivness of our approach on publicly available datasets.
\textbf{CoMA} \cite{CoMA} contains 144 dynamic sequences of 20,466 3D scans from 12 subjects.
Following \cite{CoMA}, we divide the scans into training and test sets by selecting consecutive 10 frames from every 100 frames as test samples. For evaluation purposes, we consider the facial expression of the first frame of each sequence as \emph{neutral}. Additionally, the average of the frames of the specific subject is served as the \emph{neutral} scan in the test set.
\textbf{FaceScape} \cite{FaceScape} contains static 3D scans of 847 subjects
and each subject performs 20 facial expressions (including \emph{neutral}).
We follow \cite{gu2023adversarial}
and select 30\% of the subjects randomly as the test set and the rest are used for training. Any scans of the IDs in test set is not demonstrated during training.
\textbf{D3DFACS} \cite{flame,D3DFACS} contains 519 dynamic sequences of 46,903 scans from 10 subjects and this dataset is adopted to demonstrate the feasibility of training from combined datasets.

\subsection{Implementation Details}

\textbf{Data Preprocessing}.
Our method is developed under the assumption that all scans are registered into the same topology, and correspondence learning on heterogeneous scans is beyond the scope of this paper. When training on combined datasets, all heterogeneous scans are registered into the FLAME topology. Rigid pose estimation is performed for better alignment, and we do not consider pose variations. After registration, 5\% of training scans with the lowest quality are dropped by performing PCA. The normalization parameters are pre-computed on the remaining training scans.

\textbf{Training Setups}. 
We train the network using AdamW optimizer \cite{AdamW} with common weight decay $1\times10^{-5}$ and batch size of $32$, \ie{}, $8$ IDs are selected and $4$ scans are drawn from each ID.
The network is trained for 100 epochs without data augmentation. 
Notably, $\script{L}_{mi}$ is not adopted in FaceScape ($\lambda_{mi}=0$) due to its relatively strong regularization effect resulting from a large-scale number of identity (847) compared to the latent space dimension (64).
%Detailed training setups are listed in the supplementary materials.

\newcommand{\meshvi}[2][i]{{#2}(#1)}
\newcommand{\vi}{v}
\subsection{Evaluation Metrics}
We adopt Average Vertex Distance (AVD) and the evaluation metrics used in
\cite{jiang2019disentangled,zhang2020learning,olivier2023facetunegan,gu2023adversarial}
for a fair comparison.
The AVD between the reconstructed mesh $\xrec{}$ and target mesh $X$ is defined as:
$\mathrm{AVD}(X,\xrec{})=\frac{1}{|V|}\sum_{\vi=1}^{|V|}{\lVert \meshvi[\vi{}]{X} - \meshvi[\vi{}]{\xrec{}} \rVert}_{2},$
where $|V|$ denotes the number of vertexes, and $\meshvi[\vi{}]{X}$ denotes the $\vi{}$-th vertex.% of the mesh.

\textbf{Reconstruction.}
The reconstruction error measures information preserved in the generative model, defined as:
\begin{equation}
    E_{avd}=\mathrm{AVD}(X,\xrec{}).
\end{equation}

\textbf{Disentanglement.}
The identity disentanglement metric measures the compactness of the neutralized face $\setof{\xneu{}}$ by standard deviations.
This metric is evaluated on all generated neutral meshes $\setof{\xneu{}}$ of the subject as:
\begin{equation}
    \begin{split}
    E_{id}  &=
    \mathrm{std}( {\lVert \meshvi[\vi{}]{\xneu{}}- \meshvi[\vi{}]{X^{mean}} \rVert}_{2} ), \\
    \end{split}
\end{equation}
where $\mathrm{std}()$ denotes standard deviation and $X^{mean}$ denotes the average over all generated neutral faces $\setof{\xneu{}}$ belonging to currently evaluated subject.
A similar disentanglement metric for expression $E_{exp}$ is defined analogically on all generated samples of the same expression after identity removal.

\textbf{Neutralization.}
The neutralization error measures the quality of the neutralized face $\xneu{}$ by calculating the distance between the ground-truth $X^{neu}$ and the neutralized face,
which is defined on each sample as:
\begin{equation}
    \begin{split}
      E_{neu}=\mathrm{AVD}(X^{neu}, \xneu{}).
    \end{split}
\end{equation}

\subsection{Quantitative Comparison}
We first compare the proposed \ShockingName{} with recent disentanglement methods on the FaceScape and CoMA datasets, including \DR{} \cite{jiang2019disentangled}, \DIMeshEnc{} \cite{zhang2020learning}, \IBVAE{} \cite{sun2022information}, \FTG{} \cite{olivier2023facetunegan} and \AFD{} \cite{gu2023adversarial} 
The quantitative results of reconstruction, disentanglement, and neutralization are shown in Table~\ref{tab:dataset_separate}. \FTG{} is marked in gray because they employ a different split strategy, using only a portion of scans in CoMA and reserving only 10\% of scans for evaluation in FaceScape. On CoMA, the neutralization error $E_{neu}$ of IBVAE and FED is marked in gray due to test set leakage caused by identical IDs in the training and test sets, coupled with learning neutralization through reconstruction with labeled neutral scans.

When comparing on FaceScape, our method outperforms both \AFD{} and \AFD{} without expression labels in terms of reconstruction, disentanglement, and neutralization. Notably, our method achieves a 17.4\% improvement in neutralization compared to \AFD{}, even without the advantage of expression labels. On CoMA, we observe that the compared methods, IBVAE and FED, significantly benefit from information leakage, resulting in superior neutralization performance. Comparing between FED and FED without expression labels on CoMA, 
we can see that without the aid of expression labels, the neutralization error $E_{neu}$ of \AFD{} increases tremendously (23$\times$).
%Further demonstrations of the information leakage are provided in the supplementary materials. 
Our method achieves the best reconstruction quality and disentanglement across all approaches and achieves a 47.5\% improvement in neutralization compared to its counterpart (\AFD{} without expression labels). The comparison clearly validates the effectiveness of the proposed \ShockingName{}.

\begin{table}[]
  \caption{Results achieved by training on combined datasets.}
  \label{tab:training_on_md}
  \resizebox*{\linewidth}{!}{
    \begin{tabular}{cccccccccc}
        \toprule
       \multirow{2}{*}{Dataset} & \multirow{2}{*}{w/ C.D.} & \multicolumn{2}{c}{$E_{avd}$} & \multicolumn{2}{c}{$E_{id}$} & \multicolumn{2}{c}{$E_{exp}$} & \multicolumn{2}{c}{$E_{neu}$} \\
                                &                          &      mean\lowb & median\lowb  &      mean\lowb & median\lowb &     mean\lowb & median\lowb   &   mean\lowb & median\lowb  \\
        \midrule
        \multirow{2}{*}{FaceScape*} & \xmark &     \rpfloat[3]{0.890}
                                             &     \rpfloat[3]{0.806}
                                             & \bf \rpfloat[3]{0.496}   
                                             & \bf \rpfloat[3]{0.479} 
                                             &     \rpfloat[3]{0.516} 
                                             &     \rpfloat[3]{0.517} 
                                             &     \rpfloat[3]{1.464}    
                                             &     \rpfloat[3]{1.375}    \\ % WSDF-AAAI/AE_nl2-F2_rt-nb1-jac10_gt100
                                    & \cmark & \bf \rpfloat[3]{0.841}    
                                             & \bf \rpfloat[3]{0.779}
                                             &     \rpfloat[3]{0.513}   
                                             &     \rpfloat[3]{0.492} 
                                             & \bf \rpfloat[3]{0.391} 
                                             & \bf \rpfloat[3]{0.392} 
                                             & \bf \rpfloat[3]{1.430}    
                                             & \bf \rpfloat[3]{1.347}    \\ % WSDF-AAAI/AE_nl2-F2_rt-CDV_F2-nb1-jac10_gt100
        \midrule
        \multirow{2}{*}{CoMA}       & \xmark &     \rpfloat[3]{0.480}
                                             &     \rpfloat[3]{0.409}
                                             &  \bf\rpfloat[3]{0.005}
                                             &  \bf\rpfloat[3]{0.002}
                                             & -
                                             & -
                                             &     \rpfloat[3]{0.651}  
                                             &     \rpfloat[3]{0.697}    \\ % C3 C-L3-neu10-jac10-mi10-rerun1
                                    & \cmark &  \bf\rpfloat[3]{0.449}    
                                             &  \bf\rpfloat[3]{0.401}
                                             &     \rpfloat[3]{0.023}   
                                             &     \rpfloat[3]{0.012} 
                                             & -
                                             & -
                                             &  \bf\rpfloat[3]{0.618}
                                             &  \bf\rpfloat[3]{0.648}    \\ % C3 C-L3-D1_5-nb10-jac10-mi10
        \bottomrule
    \end{tabular}
  }
\end{table}

% \subsection{Training on Combined Datasets}
\textbf{Training on Combined Datasets}.
We assess the generalizability of training our network on combined datasets. The baseline models are individually trained on the CoMA and FaceScape* datasets (registered in FLAME topology). Subsequently, we train randomly initialized models on combined datasets (with auxiliary training data from D3DFACS) using the same setups.

As shown in Table~\ref{tab:training_on_md}, training on combined datasets leads to improvements in reconstruction and neutralization for both datasets. Additionally, it enhances expression disentanglement ($E_{exp}$) due to the increased diversity of expressions, indicating improved generalizability. We attribute the slight reduction in identity disentanglement ($E_{id}$) to the limited representation space (4 dimensions for CoMA, 64 dimensions for FaceScape*) compared to the expansive corpus of identities covered by the newly introduced D3DFACS dataset. This phenomenon is particularly evident in the CoMA dataset, where the disparity between the data is notably pronounced.

\subsection{Qualitative Evaluation}

\begin{figure}[]
    \centering
    \includegraphics[width=0.8\linewidth]{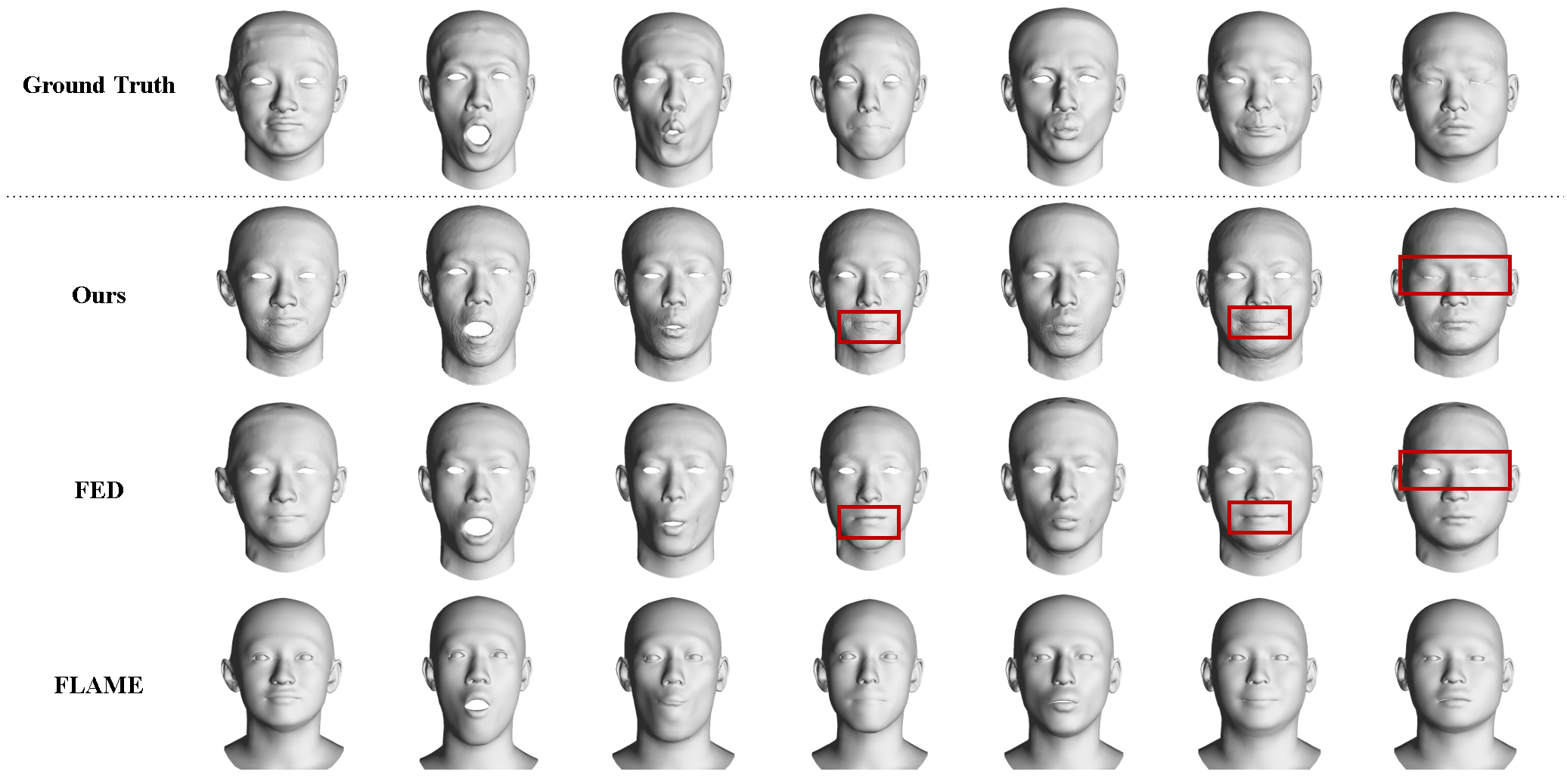}
    \caption{Qualitative comparison on FaceScape. Zoom in for a better view.}
    \label{fig:compare_recon}
\end{figure}

\textbf{Reconstruction.}
We experimentally demonstrate that our model can faithfully reconstruct face scans with unseen IDs.
To compare with FED, we directly reconstruct the facial scans in FaceScape test set with the latent codes predicted by the encoders.
As illustrated in Fig.~\ref{fig:compare_recon}, compared to the counterparts, our model is more capable of preserving the figure of facial scans 
and capturing fine structures of mouth and around the eyes. 

\begin{figure}[t]
    \centering
    \includegraphics[width=0.8\linewidth]{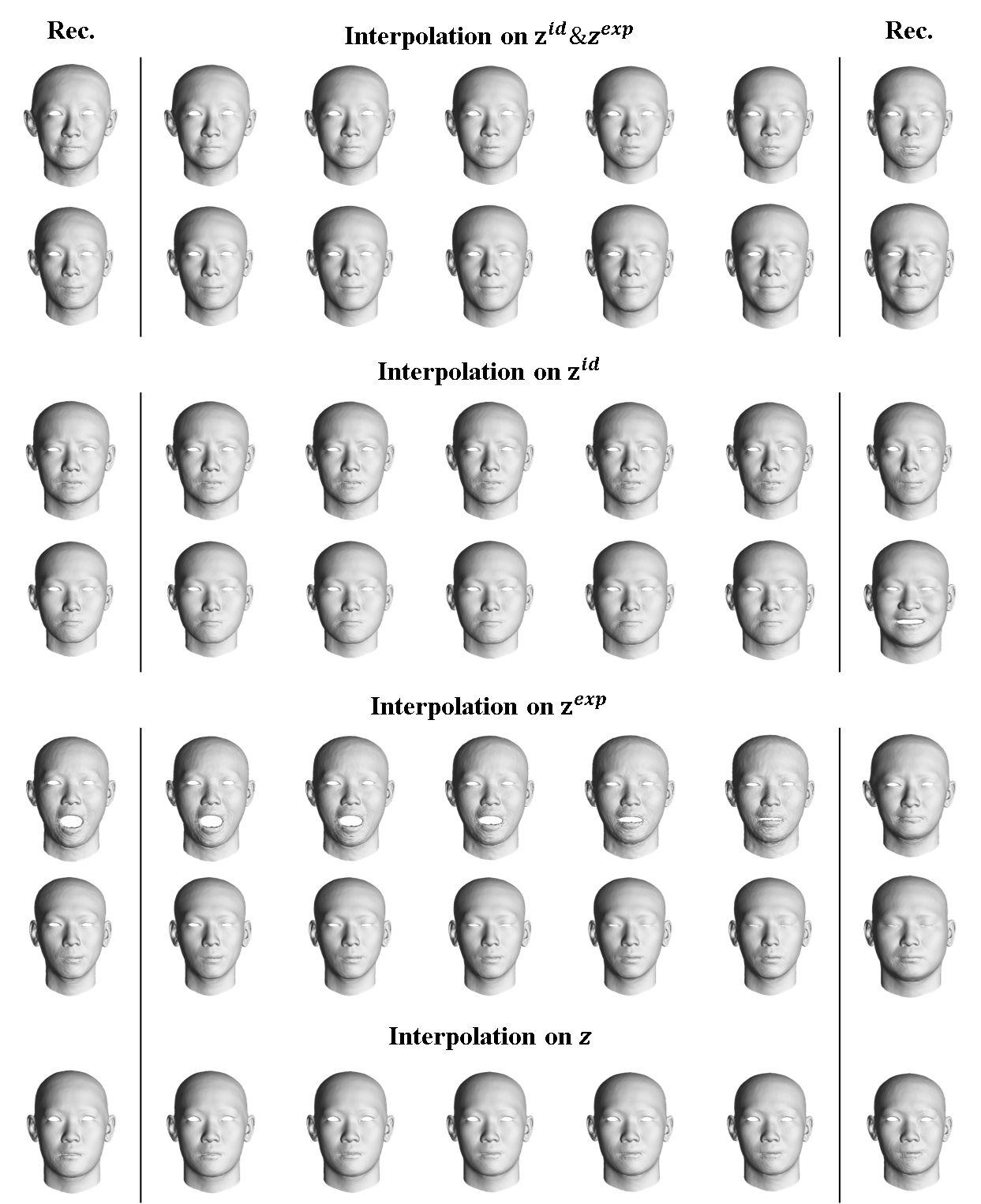}
    \caption{Interpolation results achieved on FaceScape.}
    \label{fig:visual_interpolate}
\end{figure}

\textbf{Interpolation.}
We demonstrate that the learned latent spaces are structured and disentangled by performing interpolation.
To be specific, we randomly select pairwise scans in test set (the ID is not observed during training) and calculate their latent codes, $\codei$ and $\codee$, by our encoders.
Linear interpolations are conducted jointly ($\codei$ \& $\codee$) or exclusively (using only $\codei$ or $\codee$).
It is important to note that interpolation solely on $\codei$ results in identity-swapping and interpolation solely on $\codee$ leads to expression transfer.
As illustrated in Fig.~\ref{fig:visual_interpolate}, performing linear interpolations on $\codei$ and $\codee$ simultaneously generates plausible facial geometries, and interpolation on a single factor (either $\codei$ or $\codee$) generate semantically meaningful variations.
This implies that the generation is controlled by decoupled factors that are semantically aligned.

\begin{figure}[]
    \centering
    \includegraphics[width=0.75\linewidth]{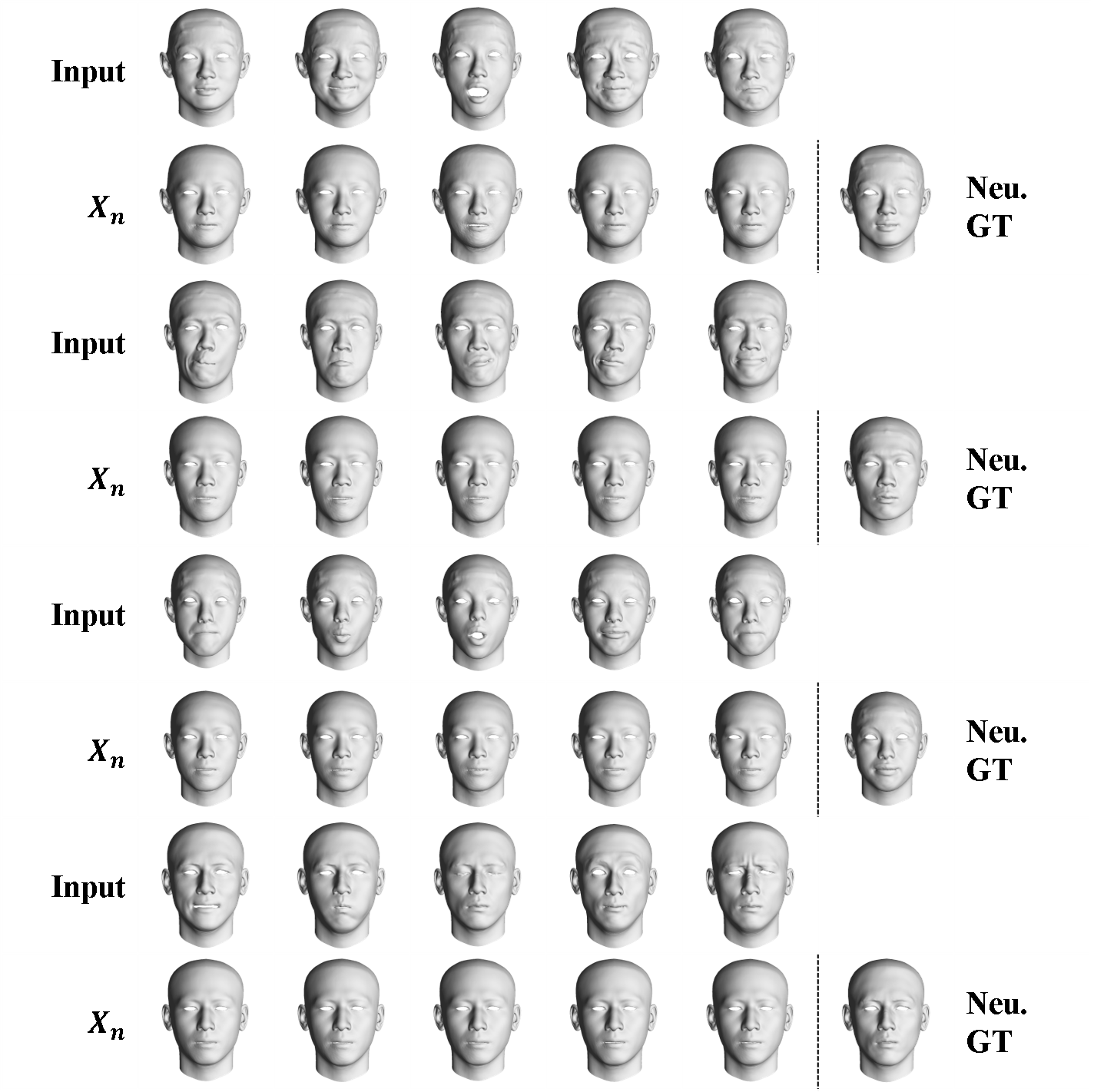}
    \caption{Neutralization of unseen scans on FaceScape. Each row indicates one individual.}
    \label{fig:visual_neutralization}
\end{figure}

\textbf{Neutralization.}
We evaluate the disentanglement performance by probing information preserved in $\codei$.
To evaluate this, we select scans with identity and expression changes from the test set (whose IDs and neutralized scans are not seen during training)
and calculate their latent codes, $\codei$ and $\codee$, using our encoders.
We then generate neutralized faces $\xneu{}$ by Eq.~\ref{eq:decode} %$using $\script{G}\circ\script{R}(\codei,\veczero)$
and compare with ground-truth scans. 
As illustrated in Fig.~\ref{fig:visual_neutralization},
the generated neutralized faces, using only $\codei$, faithfully represent identity-specified geometries and remain consistent among different scans of the same individual,
suggesting that the majority of identity-relevant information is preserved in $\codei$ without nuisance information.

\begin{table}[]
  \caption{Ablation study on FaceScape.}
  \label{tab:ablation_on_fs}
  \resizebox*{\linewidth}{!}{
    \begin{tabular}{ccccccccc}
        \toprule
        \multirow{2}{*}{Designs} & \multicolumn{2}{c}{$E_{avd}$} & \multicolumn{2}{c}{$E_{id}$} & \multicolumn{2}{c}{$E_{exp}$} & \multicolumn{2}{c}{$E_{neu}$} \\
                                  &    mean\lowb & median\lowb    &      mean\lowb & median\lowb &    mean\lowb & median\lowb    &    mean\lowb & median\lowb    \\
        \midrule
            baseline            & \bf{\rpfloat[3]{0.921}}  
                                & \bf{\rpfloat[3]{0.820}}
                                &    {\rpfloat[3]{0.994}} 
                                &    {\rpfloat[3]{0.988}}
                                &    {\rpfloat[3]{1.408}}      
                                &    {\rpfloat[3]{1.400}} 
                                &    {\rpfloat[3]{2.882}}       
                                &    {\rpfloat[3]{2.725}} \\ % C3 F-L1-baseline FaceScape:100% @1407*100
         + neu. bank            &    {\rpfloat[3]{1.052}}  
                                &    {\rpfloat[3]{0.955}}
                                & \bf{\rpfloat[3]{0.329}} 
                                & \bf{\rpfloat[3]{0.314}}
                                &    {\rpfloat[3]{0.956}}      
                                &    {\rpfloat[3]{0.942}} 
                                &    {\rpfloat[3]{1.597}}       
                                &    {\rpfloat[3]{1.498}} \\ % C3 F-L1-nb1 FaceScape:100% @1407*100
       + jac. loss              &    {\rpfloat[3]{1.110}}  
                                &    {\rpfloat[3]{0.998}}
                                &    {\rpfloat[3]{0.366}} 
                                &    {\rpfloat[3]{0.353}}
                                & \bf{\rpfloat[3]{0.466}}      
                                & \bf{\rpfloat[3]{0.479}} 
                                & \bf{\rpfloat[3]{1.590}}       
                                & \bf{\rpfloat[3]{1.485}} \\ % C3 F-L1-nb1-jac10 FaceScape:100% @1407*100
        \bottomrule
    \end{tabular}
  }
\end{table}
\begin{table}[]
  \caption{Ablation study results on CoMA.}
  \label{tab:ablation_on_coma}
  \resizebox*{\linewidth}{!}{
    \begin{tabular}{ccccccc}
        \toprule
        \multirow{2}{*}{Designs} & \multicolumn{2}{c}{$E_{avd}$} & \multicolumn{2}{c}{$E_{id}$} & \multicolumn{2}{c}{$E_{neu}$} \\
                                 &      mean\lowb & median\lowb  &      mean\lowb & median\lowb &     mean\lowb & median\lowb  \\
    \midrule
        baseline       & \bf{\rpfloat[3]{0.342}} 
                       & \bf{\rpfloat[3]{0.305}}
                       &    {\rpfloat[3]{1.611}}
                       &    {\rpfloat[3]{1.643}}
                       &    {\rpfloat[3]{3.147}}
                       &    {\rpfloat[3]{2.877}}    \\ % C3 C-L3-baseline CoMA:100% @2187*100
        + neu. bank    &    {\rpfloat[3]{0.402}}      
                       &    {\rpfloat[3]{0.356}} 
                       &    {\rpfloat[3]{0.017}}
                       &    {\rpfloat[3]{0.012}}
                       &    {\rpfloat[3]{0.659}}  
                       &    {\rpfloat[3]{0.704}}    \\ % C3 C-L3-neu10 CoMA:100% @2187*100
        + jac. loss    &    {\rpfloat[3]{0.447}}
                       &    {\rpfloat[3]{0.382}} 
                       &    {\rpfloat[3]{0.020}} 
                       &    {\rpfloat[3]{0.014}}  
                       & \bf{\rpfloat[3]{0.651}}  
                       & \bf{\rpfloat[3]{0.684}}    \\ % C3 C-L3-neu10-jac10 CoMA:100% @2187*100
        + mi. loss     &    {\rpfloat[3]{0.480}}
                       &    {\rpfloat[3]{0.409}} 
                       & \bf{\rpfloat[3]{0.005}} 
                       & \bf{\rpfloat[3]{0.002}}  
                       & \bf{\rpfloat[3]{0.651}}  
                       &    {\rpfloat[3]{0.697}}    \\ % C3 C-L3-neu10-jac10-mi10-rerun1 CoMA:100% @2187*100
        \bottomrule
    \end{tabular}
  }
\end{table}

\subsection{Ablation Study}

\begin{figure}[]
    \centering
    \includegraphics[width=0.95\linewidth]{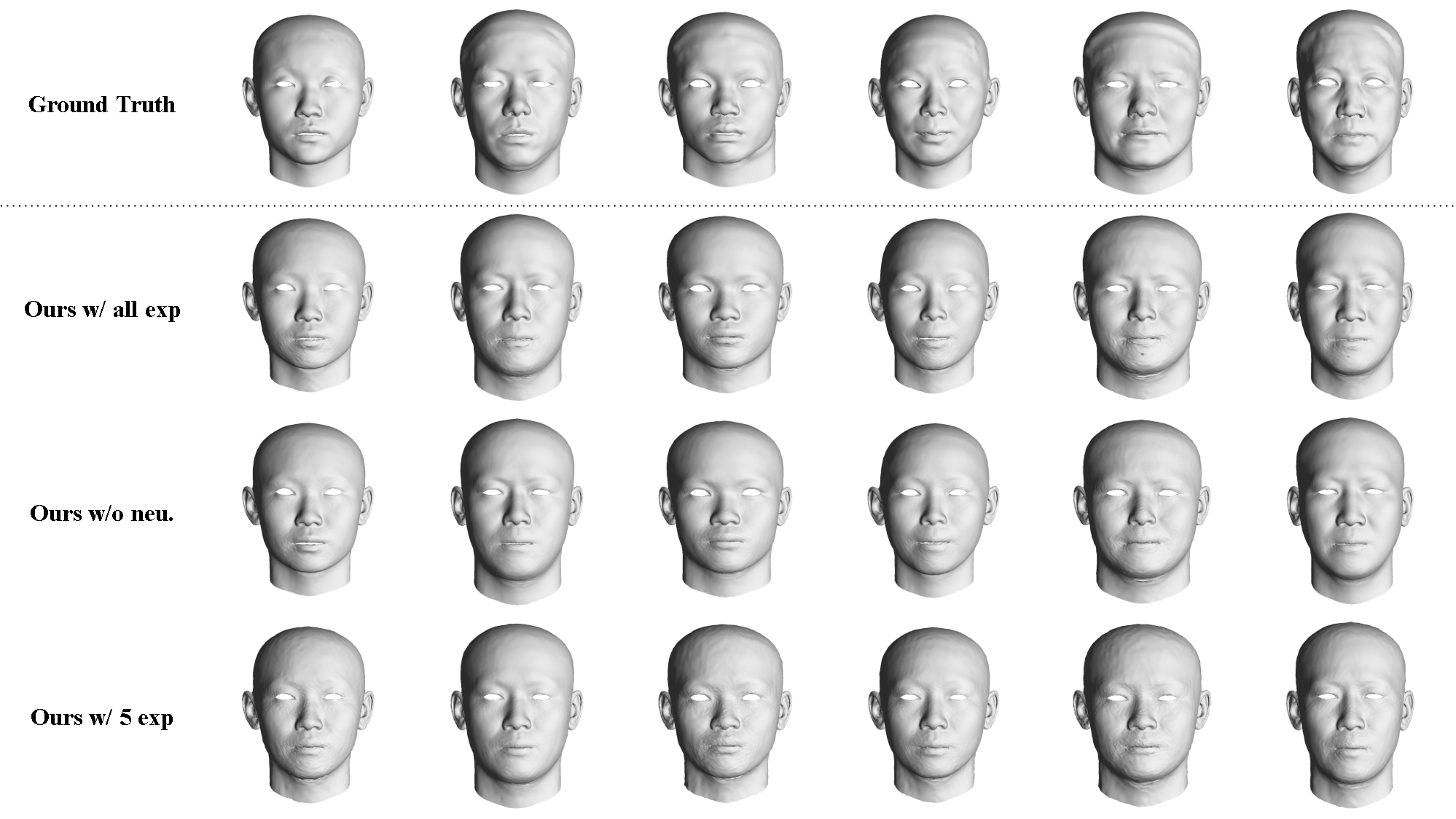}
    \caption{Visualization of Neutral Bank on FaceScape.}
    \label{fig:visual_neutral_bank}
\end{figure}

\textbf{Effect of \NBankName{}}. We investigate its necessity by training with only $\script{L}_{rec}$ and $\script{L}_{KL}$ as the baseline method, %, while keeping the same pipeline with the exception of EMA update.
no supervision is applied on $\xneu{}$.
%and thus, the degenerated solution is not prevented.
As illustrated in Table~\ref{tab:ablation_on_fs}, the baseline method demonstrates proficient reconstruction capabilities, as evidenced by its low $E_{avd}$. However, it exhibits shortcomings, as reflected in the elevated values of $E_{id}$ and $E_{neu}$. These higher values suggest that the baseline struggles to disentangle underlying controlling factors, resulting in the inability to generate convincing neutralized faces and maintain consistency in identity representation. %, which is a degenerated solution to disentanglement.
% Comparatively, the disentanglement metric $E_{id}$, $E_{exp}$ and the neutralization error $E_{neu}$ decrease significantly by adding \NBankName{} (+ neu. bank).
Comparatively, $E_{id}$, $E_{exp}$ and $E_{neu}$ decrease significantly by adding \NBankName{} (+ neu. bank).
This finding supports that the proposed \NBankName{} is crucial for identity-consistency and disentanglement learning. 
Similarly, this necessity is also evident when training on CoMA as show in Table~\ref{tab:ablation_on_coma} (baseline \textit{v.s.} + neu. bank).

We also qualitatively evaluate Neutral Bank by comparing with ground-truth neural scans on FaceScape. As shown in Fig. ~\ref{fig:visual_neutral_bank}, when training with all expressions (ours w/ all exp), the EMA updated Neutral Bank preserves identity-specified geometries. 
The model could still capture identity-relevant geometries when neutral scans are excluded from training (ours w/o neu.), and even when only 5 expressions (ours w/ 5 exp) are observed during training.
This suggests the robustness of \NBankName{} to the existence of scans with neutral expression and the diversity of expressions.

\textbf{Effect of \JLossName{}}. 
Incorporating $\mathcal{L}_{jac}$ to our framework (with a neutralization bank), as presented in Table~\ref{tab:ablation_on_fs}, has yielded noteworthy outcomes. Notably, the disentanglement metric $E_{exp}$ has seen a substantial reduction of 51.2\%, indicating a marked improvement. There is a slight decrease in $E_{neu}$, and other metrics show marginal impact, attributable to the delicate balance struck among multiple losses.
This outcome suggests that the inclusion of the second-order loss, $\mathcal{L}_{jac}$, has proven effective in prompting the expression branch to discern and isolate nuisance information, thereby enhancing overall disentanglement. Additionally, on CoMA, the introduction of $\mathcal{L}_{mi}$ has facilitated a more favorable trade-off between the disentangled factors.

\begin{figure}[]
    \centering
    \includegraphics[width=0.75\linewidth]{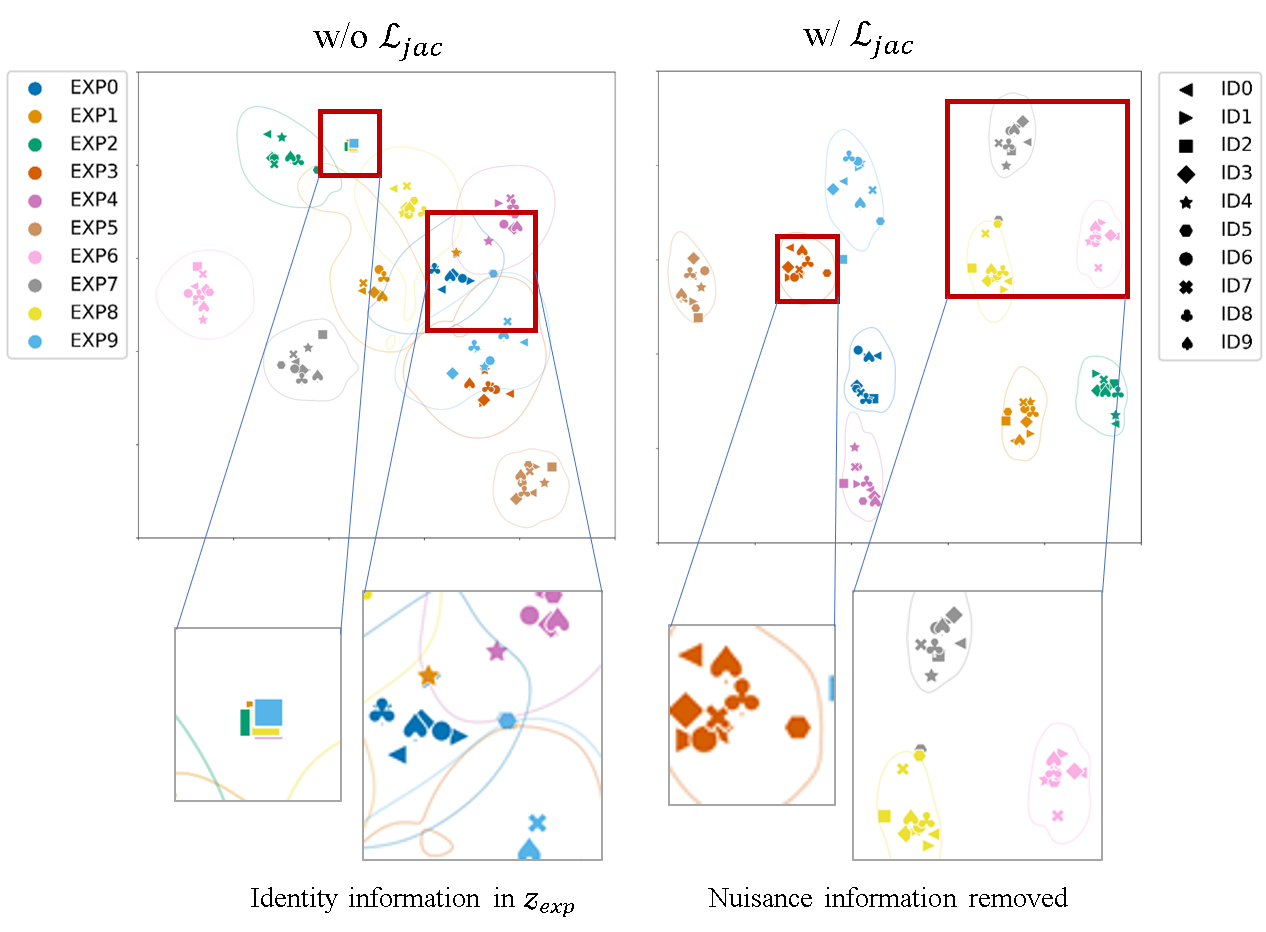}
    \caption{t-SNE of $z_{exp}$ on FaceScape test set. Shaped by IDs.}
    \label{fig:visual_tsne}
\end{figure}

We further performs t-SNE analysis on the predicted $z_{exp}$ from FaceScape test set.
As illustrated in Fig.~\ref{fig:visual_tsne}, the expression representations $z_{exp}$ exhibit identity-based clustering without using $\script{L}_{jac}$, indicating the presence of irrelevant information.
In contrast, employing $\mathcal{L}_{jac}$ facilitates eliminating such extraneous information, where the clustering of $z_{exp}$ is only associated with expression categories.

\subsection{Application}
\begin{figure}[]
  \centering
  \begin{subfigure}{0.4\textwidth}
      \includegraphics[width=\linewidth]{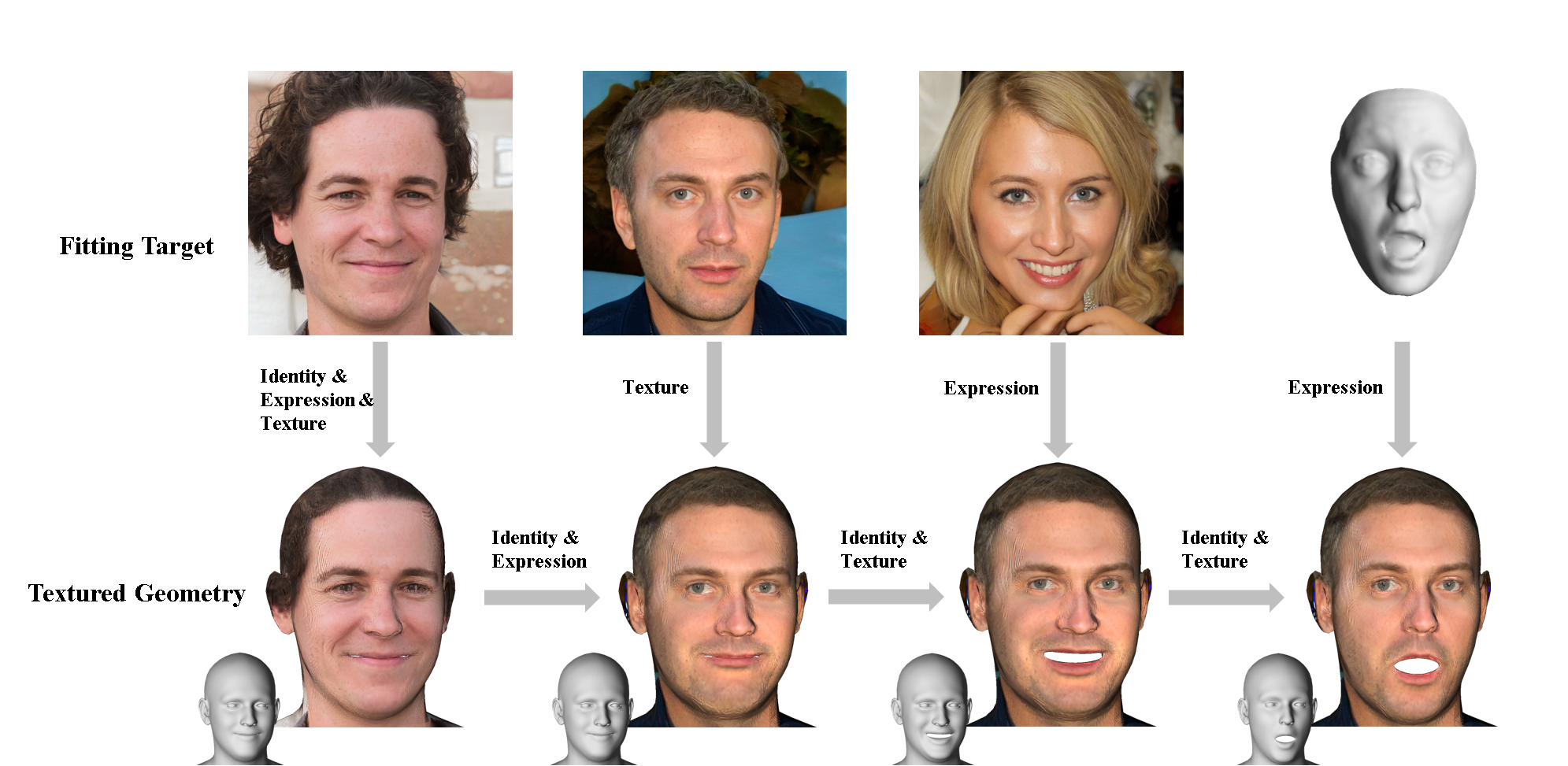}
      \caption{
        Reconstruction and editing. %First column: 3D face reconstruction result by fitting 2D image. Second column: texture transfer from 2D image. Third column: expression transfer from 2D image. Last column: expression transfer from fitted 3D scan.
     }
      \label{fig:vis_app_rec}
  \end{subfigure}
  \\
  \begin{subfigure}{0.4\textwidth}
     \includegraphics[width=\linewidth]{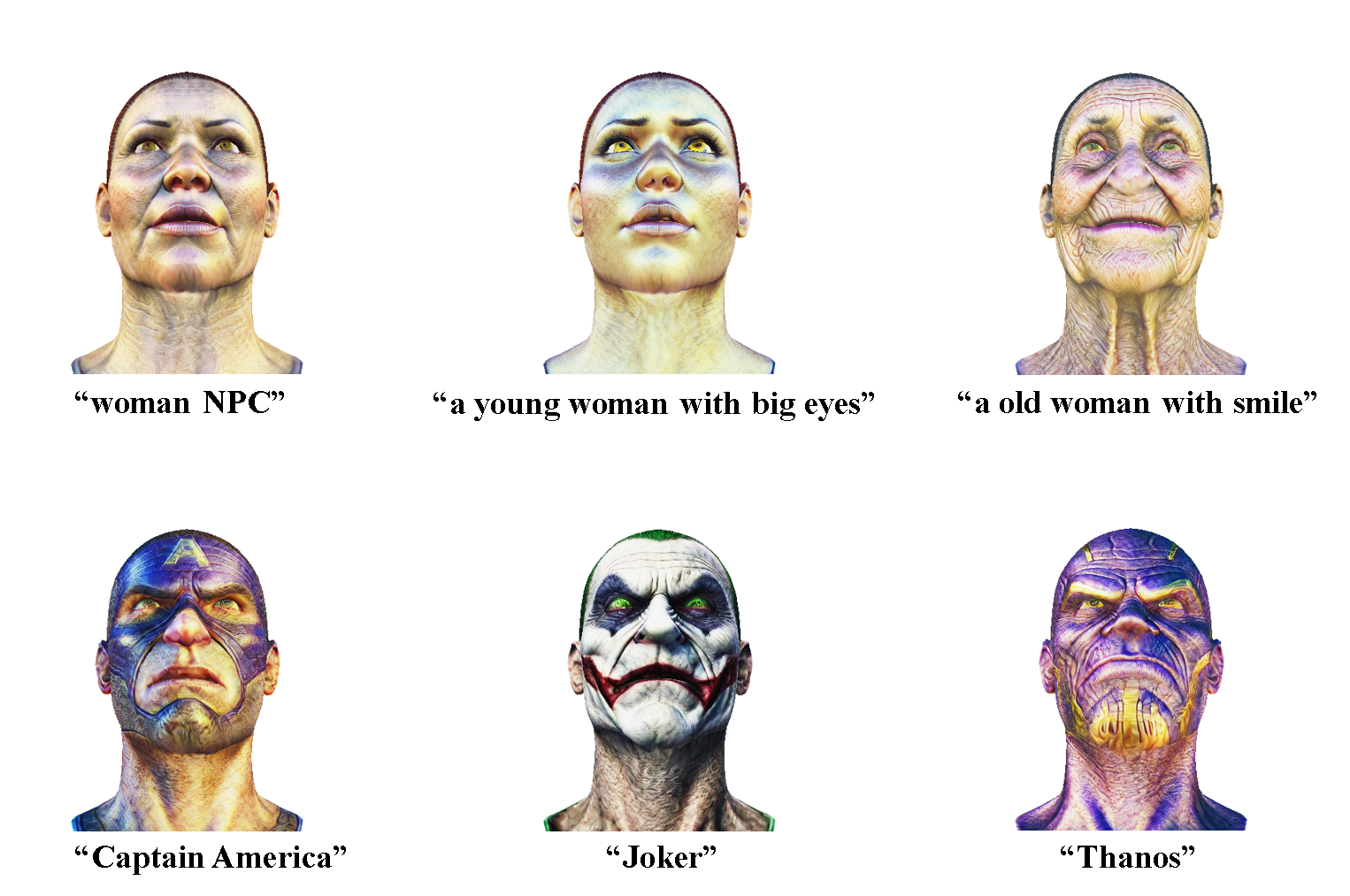}
      \caption{
        Text-to-3D generation.
     }
      \label{fig:vis_app_t3d}
  \end{subfigure}
  
  \caption{
    Examples of downstream applications.
 }
  \label{fig:vis_application}
\end{figure}

The learned generative 3D face model achieves controllability via disentanglement learning %and generalizabilty by training on combined datasets,
and it can be applied to a number of downstream tasks.
In Fig.~\ref{fig:vis_application}(a), we provide 3D face reconstruction (first column), texture transfer (second column) and expression transfer (third column) results by fitting 2D images (generated by StyleGAN2 \cite{StyleGAN2-ADA}), and the expression transfer result (last column) by fitting 3D scan (from BU-3DFE \cite{BU-3DFE}).
% In particular, we fit the learned 3D face model to 2D images with landmark loss and image reconstruction loss along with constraints on symmetry of texture map, shading coefficients, norm of rigid pose and norm of latent codes.
% In particular, we fit the learned 3D face model to 2D images with landmark loss and image reconstruction loss along with constraints on symmetry of texture map and norm on latent codes.
Texture transfer and expression transfer are achieved by swapping the fitted texture maps and expression latent codes.
% As for fitting 3D scans, besides the constraint on the latent codes and rigid pose, the 3D face model is fitted with truncated chamfer distance and landmark loss.
Similarly to 2D expression transfer, the expression is transferred by exchanging expression latent codes.

Fig.~\ref{fig:vis_application}(b) provides text-to-3D generation by utilizing our decoding components, the pre-trained diffusion model~\cite{stable_diffusion} and SDS loss~\cite{sds_loss}.
Our pipeline draws inspiration from Fantasia3D~\cite{fantasia3d} and utilizes a mesh-based differentiable renderer~\cite{nvdiffrast} to simultaneously optimize parameterized geometry and texture. 
Thanks to our mesh-based 3D representation, the generation pipeline is efficient and it takes only 15 minutes to optimize a 3D face on a single RTX 3090.

%%%%%%%%%%%%%%%%%%%%%%%%%%%%%%%%%%%%%%%%%%%%%%%%%%%%%%%%%%%%%%%%%%%%%%%%%%%%%%%%
\section{CONCLUSIONS}
% \subsection{Conclusions}

This paper presents a method for controllable 3D face modeling with weakly-supervised training. The Neutral Bank module utilizes identity labels to learn pseudo ground-truth, enforcing identity-consistency along with information preservation through an auxiliary loss function. Additionally, a label-free second-order loss function is designed to further enhance disentanglement by imposing regularization on the latent expression space. Experimental results demonstrate that the proposed \ShockingName{} is effective in learning a controllable 3D face model. This study also validates the potential of learning 3D face models from multiple collections.%, showing improved generalizability.

% \subsection{Discussions}

% scalability

% social impact

%%%%%%%%%%%%%%%%%%%%%%%%%%%%%%%%%%%%%%%%%%%%%%%%%%%%%%%%%%%%%%%%%%%%%%%%%%%%%%%%
\section{ACKNOWLEDGMENTS}

This work is partly supported by the National Key R\&D Program of China (No. 2022ZD0161902), the National Natural Science Foundation of China (No. 62176012, 62202031), and the Beijing Natural Science Foundation (No. 4222049).

%%%%%%%%%%%%%%%%%%%%%%%%%%%%%%%%%%%%%%%%%%%%%%%%%%%%%%%%%%%%%%%%%%%%%%%%%%%%%%%%

{\small
\bibliographystyle{ieee}
\bibliography{
main_bib,
bib_fm,   % face model
bib_de,   % disentanglement
bib_ds,   % dataset
bib_dm,   % diffusion model
bib_misc  % misc
}

\begin{thebibliography}{10}\itemsep=-1pt

\bibitem{Amberg2008:ExpressionInvariant}
B.~Amberg, R.~Knothe, and T.~Vetter.
\newblock Expression invariant 3d face recognition with a morphable model.
\newblock In {\em 2008 8th IEEE International Conference on Automatic Face \& Gesture Recognition}, pages 1--6, 2008.

\bibitem{CompositionalVAEs}
T.~Bagautdinov, C.~Wu, J.~Saragih, P.~Fua, and Y.~Sheikh.
\newblock Modeling facial geometry using compositional vaes.
\newblock In {\em 2018 IEEE/CVF Conference on Computer Vision and Pattern Recognition}, pages 3877--3886, 2018.

\bibitem{Bahri2021:SMF}
M.~Bahri, E.~O'~Sullivan, S.~Gong, F.~Liu, X.~Liu, M.~M. Bronstein, and S.~Zafeiriou.
\newblock Shape my face: Registering 3d face scans by surface-to-surface translation.
\newblock {\em International Journal of Computer Vision}, 129(9):2680--2713, Sep 2021.

\bibitem{3DMM:99}
V.~Blanz and T.~Vetter.
\newblock A morphable model for the synthesis of 3d faces.
\newblock In {\em Proceedings of the 26th Annual Conference on Computer Graphics and Interactive Techniques}, SIGGRAPH '99, page 187–194, USA, 1999. ACM Press/Addison-Wesley Publishing Co.

\bibitem{3DMM:03}
V.~Blanz and T.~Vetter.
\newblock Face recognition based on fitting a 3d morphable model.
\newblock {\em IEEE Transactions on Pattern Analysis and Machine Intelligence}, 25(9):1063--1074, 2003.

\bibitem{LSFM}
J.~Booth, A.~Roussos, S.~Zafeiriou, A.~Ponniah, and D.~Dunaway.
\newblock A 3d morphable model learnt from 10,000 faces.
\newblock In {\em 2016 IEEE Conference on Computer Vision and Pattern Recognition (CVPR)}, pages 5543--5552, 2016.

\bibitem{Neural3DMM}
G.~Bouritsas, S.~Bokhnyak, S.~Ploumpis, S.~Zafeiriou, and M.~Bronstein.
\newblock Neural 3d morphable models: Spiral convolutional networks for 3d shape representation learning and generation.
\newblock In {\em 2019 IEEE/CVF International Conference on Computer Vision (ICCV)}, pages 7212--7221, 2019.

\bibitem{FaceWarehouse}
C.~Cao, Y.~Weng, S.~Zhou, Y.~Tong, and K.~Zhou.
\newblock Facewarehouse: A 3d facial expression database for visual computing.
\newblock {\em IEEE Transactions on Visualization and Computer Graphics}, 20(3):413--425, 2014.

\bibitem{fantasia3d}
R.~Chen, Y.~Chen, N.~Jiao, and K.~Jia.
\newblock Fantasia3d: Disentangling geometry and appearance for high-quality text-to-3d content creation.
\newblock In {\em Proceedings of the IEEE/CVF International Conference on Computer Vision (ICCV)}, October 2023.

\bibitem{chen2018:TCVAE}
R.~T.~Q. Chen, X.~Li, R.~B. Grosse, and D.~K. Duvenaud.
\newblock Isolating sources of disentanglement in variational autoencoders.
\newblock In S.~Bengio, H.~Wallach, H.~Larochelle, K.~Grauman, N.~Cesa-Bianchi, and R.~Garnett, editors, {\em Advances in Neural Information Processing Systems}, volume~31. Curran Associates, Inc., 2018.

\bibitem{elu}
D.~Clevert, T.~Unterthiner, and S.~Hochreiter.
\newblock Fast and accurate deep network learning by exponential linear units (elus).
\newblock In Y.~Bengio and Y.~LeCun, editors, {\em 4th International Conference on Learning Representations, {ICLR} 2016, San Juan, Puerto Rico, May 2-4, 2016, Conference Track Proceedings}, 2016.

\bibitem{D3DFACS}
D.~Cosker, E.~Krumhuber, and A.~Hilton.
\newblock A facs valid 3d dynamic action unit database with applications to 3d dynamic morphable facial modeling.
\newblock In {\em 2011 International Conference on Computer Vision}, pages 2296--2303, 2011.

\bibitem{Danecek2022:EMOCA}
R.~Danecek, M.~J. Black, and T.~Bolkart.
\newblock {EMOCA}: {E}motion driven monocular face capture and animation.
\newblock In {\em Conference on Computer Vision and Pattern Recognition (CVPR)}, pages 20311--20322, 2022.

\bibitem{Deng2019:Accurate}
Y.~Deng, J.~Yang, S.~Xu, D.~Chen, Y.~Jia, and X.~Tong.
\newblock Accurate 3d face reconstruction with weakly-supervised learning: From single image to image set.
\newblock In {\em 2019 IEEE/CVF Conference on Computer Vision and Pattern Recognition Workshops (CVPRW)}, pages 285--295, 2019.

\bibitem{Egger2020:3DMMPPF}
B.~Egger, W.~A.~P. Smith, A.~Tewari, S.~Wuhrer, M.~Zollhoefer, T.~Beeler, F.~Bernard, T.~Bolkart, A.~Kortylewski, S.~Romdhani, C.~Theobalt, V.~Blanz, and T.~Vetter.
\newblock 3d morphable face models—past, present, and future.
\newblock {\em ACM Trans. Graph.}, 39(5), jun 2020.

\bibitem{Galanakis2023:3DMMRF}
S.~Galanakis, B.~Gecer, A.~Lattas, and S.~Zafeiriou.
\newblock 3dmm-rf: Convolutional radiance fields for 3d face modeling.
\newblock In {\em Proceedings of the IEEE/CVF Winter Conference on Applications of Computer Vision (WACV)}, pages 3536--3547, January 2023.

\bibitem{BFM2017}
T.~Gerig, A.~Morel-Forster, C.~Blumer, B.~Egger, M.~Luthi, S.~Schoenborn, and T.~Vetter.
\newblock Morphable face models - an open framework.
\newblock In {\em 2018 13th IEEE International Conference on Automatic Face \& Gesture Recognition (FG 2018)}, pages 75--82, 2018.

\bibitem{Gong2019:Spiral}
S.~Gong, L.~Chen, M.~Bronstein, and S.~Zafeiriou.
\newblock Spiralnet++: A fast and highly efficient mesh convolution operator.
\newblock In {\em 2019 IEEE/CVF International Conference on Computer Vision Workshop (ICCVW)}, pages 4141--4148, 2019.

\bibitem{gu2023adversarial}
Y.~Gu, N.~Pears, and H.~Sun.
\newblock Adversarial 3d face disentanglement of identity and expression.
\newblock In {\em 2023 IEEE 17th International Conference on Automatic Face and Gesture Recognition (FG)}, pages 1--7, 2023.

\bibitem{Higgins2017:betaVAE}
I.~Higgins, L.~Matthey, A.~Pal, C.~P. Burgess, X.~Glorot, M.~M. Botvinick, S.~Mohamed, and A.~Lerchner.
\newblock beta-vae: Learning basic visual concepts with a constrained variational framework.
\newblock In {\em 5th International Conference on Learning Representations, {ICLR} 2017, Toulon, France, April 24-26, 2017, Conference Track Proceedings}. OpenReview.net, 2017.

\bibitem{hong2021headnerf}
Y.~Hong, B.~Peng, H.~Xiao, L.~Liu, and J.~Zhang.
\newblock Headnerf: A real-time nerf-based parametric head model.
\newblock In {\em {IEEE/CVF} Conference on Computer Vision and Pattern Recognition (CVPR)}, 2022.

\bibitem{jiang2019disentangled}
Z.-H. Jiang, Q.~Wu, K.~Chen, and J.~Zhang.
\newblock Disentangled representation learning for 3d face shape.
\newblock In {\em 2019 IEEE/CVF Conference on Computer Vision and Pattern Recognition (CVPR)}, pages 11949--11958, 2019.

\bibitem{kacem2022disentangled}
A.~Kacem, K.~Cherenkova, and D.~Aouada.
\newblock Disentangled face identity representations for joint 3d face recognition and neutralisation.
\newblock In {\em 2022 8th International Conference on Virtual Reality (ICVR)}, pages 438--443, 2022.

\bibitem{StyleGAN2-ADA}
T.~Karras, M.~Aittala, J.~Hellsten, S.~Laine, J.~Lehtinen, and T.~Aila.
\newblock Training generative adversarial networks with limited data.
\newblock In {\em Proc. NeurIPS}, 2020.

\bibitem{kwon2021:diagonal}
G.~Kwon and J.~C. Ye.
\newblock Diagonal attention and style-based gan for content-style disentanglement in image generation and translation.
\newblock In {\em Proceedings of the IEEE/CVF International Conference on Computer Vision}, pages 13980--13989, 2021.

\bibitem{nvdiffrast}
S.~Laine, J.~Hellsten, T.~Karras, Y.~Seol, J.~Lehtinen, and T.~Aila.
\newblock Modular primitives for high-performance differentiable rendering.
\newblock {\em ACM Transactions on Graphics}, 39(6), 2020.

\bibitem{flame}
T.~Li, T.~Bolkart, M.~J. Black, H.~Li, and J.~Romero.
\newblock Learning a model of facial shape and expression from 4d scans.
\newblock {\em ACM Trans. Graph.}, 36(6), nov 2017.

\bibitem{Lin2022:SDVAE}
J.~Ling, Z.~Wang, M.~Lu, Q.~Wang, C.~Qian, and F.~Xu.
\newblock Semantically disentangled variational autoencoder for modeling 3d facial details.
\newblock {\em IEEE Transactions on Visualization and Computer Graphics}, pages 1--1, 2022.

\bibitem{Liu2019:3DFace}
F.~Liu, L.~Tran, and X.~Liu.
\newblock 3d face modeling from diverse raw scan data.
\newblock In {\em 2019 IEEE/CVF International Conference on Computer Vision (ICCV)}, pages 9407--9417, 2019.

\bibitem{Liu2020:Joint}
F.~Liu, Q.~Zhao, X.~Liu, and D.~Zeng.
\newblock Joint face alignment and 3d face reconstruction with application to face recognition.
\newblock {\em IEEE Transactions on Pattern Analysis and Machine Intelligence}, 42(3):664--678, 2020.

\bibitem{Locatello2019:Challenging}
F.~Locatello, S.~Bauer, M.~Lucic, G.~Raetsch, S.~Gelly, B.~Sch{\"o}lkopf, and O.~Bachem.
\newblock Challenging common assumptions in the unsupervised learning of disentangled representations.
\newblock In {\em International Conference on Machine Learning}, pages 4114--4124, 2019.

\bibitem{AdamW}
I.~Loshchilov and F.~Hutter.
\newblock Fixing weight decay regularization in adam.
\newblock {\em CoRR}, abs/1711.05101, 2017.

\bibitem{olivier2023facetunegan}
N.~Olivier, K.~Baert, F.~Danieau, F.~Multon, and Q.~Avril.
\newblock Facetunegan: Face autoencoder for convolutional expression transfer using neural generative adversarial networks.
\newblock {\em Comput. Graph.}, 110(C):69–85, mar 2023.

\bibitem{pang2023dpe}
Y.~Pang, Y.~Zhang, W.~Quan, Y.~Fan, X.~Cun, Y.~Shan, and D.-m. Yan.
\newblock Dpe: Disentanglement of pose and expression for general video portrait editing.
\newblock {\em arXiv preprint arXiv:2301.06281}, 2023.

\bibitem{BFM2009}
P.~Paysan, R.~Knothe, B.~Amberg, S.~Romdhani, and T.~Vetter.
\newblock A 3d face model for pose and illumination invariant face recognition.
\newblock In {\em 2009 Sixth IEEE International Conference on Advanced Video and Signal Based Surveillance}, pages 296--301, 2009.

\bibitem{Ploumpis2019:Combining}
S.~Ploumpis, H.~Wang, N.~Pears, W.~A.~P. Smith, and S.~Zafeiriou.
\newblock Combining 3d morphable models: A large scale face-and-head model.
\newblock In {\em 2019 IEEE/CVF Conference on Computer Vision and Pattern Recognition (CVPR)}, pages 10926--10935, 2019.

\bibitem{sds_loss}
B.~Poole, A.~Jain, J.~T. Barron, and B.~Mildenhall.
\newblock Dreamfusion: Text-to-3d using 2d diffusion.
\newblock In {\em The Eleventh International Conference on Learning Representations, {ICLR} 2023, Kigali, Rwanda, May 1-5, 2023}. OpenReview.net, 2023.

\bibitem{R2021:Learning}
M.~B. R, A.~Tewari, H.-P. Seidel, M.~Elgharib, and C.~Theobalt.
\newblock Learning complete 3d morphable face models from images and videos.
\newblock In {\em 2021 IEEE/CVF Conference on Computer Vision and Pattern Recognition (CVPR)}, pages 3360--3370, 2021.

\bibitem{CoMA}
A.~Ranjan, T.~Bolkart, S.~Sanyal, and M.~J. Black.
\newblock Generating {3D} faces using convolutional mesh autoencoders.
\newblock In {\em European Conference on Computer Vision (ECCV)}, pages 725--741, 2018.

\bibitem{stable_diffusion}
R.~Rombach, A.~Blattmann, D.~Lorenz, P.~Esser, and B.~Ommer.
\newblock High-resolution image synthesis with latent diffusion models.
\newblock In {\em Proceedings of the IEEE/CVF Conference on Computer Vision and Pattern Recognition (CVPR)}, pages 10684--10695, June 2022.

\bibitem{Shu2020:Weakly}
R.~Shu, Y.~Chen, A.~Kumar, S.~Ermon, and B.~Poole.
\newblock Weakly supervised disentanglement with guarantees.
\newblock In {\em 8th International Conference on Learning Representations, {ICLR} 2020, Addis Ababa, Ethiopia, April 26-30, 2020}. OpenReview.net, 2020.

\bibitem{Albedo3DMM}
W.~A.~P. Smith, A.~Seck, H.~Dee, B.~Tiddeman, J.~B. Tenenbaum, and B.~Egger.
\newblock A morphable face albedo model.
\newblock In {\em 2020 IEEE/CVF Conference on Computer Vision and Pattern Recognition (CVPR)}, pages 5010--5019, 2020.

\bibitem{sun2022information}
H.~Sun, N.~Pears, and Y.~Gu.
\newblock Information bottlenecked variational autoencoder for disentangled 3d facial expression modelling.
\newblock In {\em 2022 IEEE/CVF Winter Conference on Applications of Computer Vision (WACV)}, pages 2334--2343, 2022.

\bibitem{sun2023next3d}
J.~Sun, X.~Wang, L.~Wang, X.~Li, Y.~Zhang, H.~Zhang, and Y.~Liu.
\newblock Next3d: Generative neural texture rasterization for 3d-aware head avatars.
\newblock In {\em CVPR}, 2023.

\bibitem{Tewari2019:FML}
A.~Tewari, F.~Bernard, P.~Garrido, G.~Bharaj, M.~Elgharib, H.-P. Seidel, P.~Pérez, M.~Zollhöfer, and C.~Theobalt.
\newblock Fml: Face model learning from videos.
\newblock In {\em 2019 IEEE/CVF Conference on Computer Vision and Pattern Recognition (CVPR)}, pages 10804--10814, 2019.

\bibitem{Thies2018:Face2Face}
J.~Thies, M.~Zollh\"{o}fer, M.~Stamminger, C.~Theobalt, and M.~Nie\ss{}ner.
\newblock Face2face: Real-time face capture and reenactment of rgb videos.
\newblock {\em Commun. ACM}, 62(1):96–104, dec 2018.

\bibitem{Tran2019:Towards}
L.~Tran, F.~Liu, and X.~Liu.
\newblock Towards high-fidelity nonlinear 3d face morphable model.
\newblock In {\em In Proceeding of IEEE Computer Vision and Pattern Recognition}, Long Beach, CA, June 2019.

\bibitem{Tran2018:Nonlinear}
L.~Tran and X.~Liu.
\newblock Nonlinear 3d face morphable model.
\newblock In {\em IEEE Computer Vision and Pattern Recognition (CVPR)}, Salt Lake City, UT, June 2018.

\bibitem{Tran2021:OnLearning}
L.~Tran and X.~Liu.
\newblock On learning 3d face morphable model from in-the-wild images.
\newblock {\em IEEE Trans. Pattern Anal. Mach. Intell.}, 43(1):157–171, jan 2021.

\bibitem{ulyanov2017instance}
D.~Ulyanov, A.~Vedaldi, and V.~Lempitsky.
\newblock Instance normalization: The missing ingredient for fast stylization, 2017.

\bibitem{bilinear}
D.~Vlasic, M.~Brand, H.~Pfister, and J.~Popovi\'{c}.
\newblock Face transfer with multilinear models.
\newblock {\em ACM Trans. Graph.}, 24(3):426–433, jul 2005.

\bibitem{bilinear:siggraph}
D.~Vlasic, M.~Brand, H.~Pfister, and J.~Popovic.
\newblock Face transfer with multilinear models.
\newblock In {\em ACM SIGGRAPH 2006 Courses}, SIGGRAPH '06, page 24–es, New York, NY, USA, 2006. Association for Computing Machinery.

\bibitem{Wang2022:FaceVerse}
L.~Wang, Z.~Chen, T.~Yu, C.~Ma, L.~Li, and Y.~Liu.
\newblock Faceverse: a fine-grained and detail-controllable 3d face morphable model from a hybrid dataset.
\newblock In {\em IEEE Conference on Computer Vision and Pattern Recognition (CVPR2022)}, June 2022.

\bibitem{yue2022anifacegan}
Y.~Wu, Y.~Deng, J.~Yang, F.~Wei, C.~Qifeng, and X.~Tong.
\newblock Anifacegan: Animatable 3d-aware face image generation for video avatars.
\newblock In {\em Advances in Neural Information Processing Systems}, 2022.

\bibitem{FaceScape}
H.~Yang, H.~Zhu, Y.~Wang, M.~Huang, Q.~Shen, R.~Yang, and X.~Cao.
\newblock Facescape: a large-scale high quality 3d face dataset and detailed riggable 3d face prediction.
\newblock In {\em Proceedings of the IEEE Conference on Computer Vision and Pattern Recognition (CVPR)}, 2020.

\bibitem{Yang2022:BareSkinNet}
X.~Yang and T.~Taketomi.
\newblock Bareskinnet: De-makeup and de-lighting via 3d face reconstruction.
\newblock {\em Computer Graphics Forum}, 41(7):623--634, 2022.

\bibitem{Yenamandra2021:i3DMM}
T.~Yenamandra, A.~Tewari, F.~Bernard, H.-P. Seidel, M.~Elgharib, D.~Cremers, and C.~Theobalt.
\newblock i3dmm: Deep implicit 3d morphable model of human heads.
\newblock In {\em 2021 IEEE/CVF Conference on Computer Vision and Pattern Recognition (CVPR)}, pages 12798--12808, 2021.

\bibitem{BU-3DFE}
L.~Yin, X.~Wei, Y.~Sun, J.~Wang, and M.~Rosato.
\newblock A 3d facial expression database for facial behavior research.
\newblock In {\em 7th International Conference on Automatic Face and Gesture Recognition (FGR06)}, pages 211--216, 2006.

\bibitem{Zhang2021:Learning}
W.~Zhang, X.~Ji, K.~Chen, Y.~Ding, and C.~Fan.
\newblock Learning a facial expression embedding disentangled from identity.
\newblock In {\em 2021 IEEE/CVF Conference on Computer Vision and Pattern Recognition (CVPR)}, pages 6755--6764, 2021.

\bibitem{Zhang2023:NPF}
Z.~Zhang, R.~Chen, W.~Cao, Y.~Tai, and C.~Wang.
\newblock Learning neural proto-face field for disentangled 3d face modeling in the wild.
\newblock In {\em Proceedings of the IEEE/CVF Conference on Computer Vision and Pattern Recognition (CVPR)}, pages 382--393, June 2023.

\bibitem{zhang2020learning}
Z.~Zhang, C.~Yu, H.~Li, J.~Sun, and F.~Liu.
\newblock Learning distribution independent latent representation for 3d face disentanglement.
\newblock In {\em 2020 International Conference on 3D Vision (3DV)}, pages 848--857, 2020.

\bibitem{Zheng2022:ImFace}
M.~Zheng, H.~Yang, D.~Huang, and L.~Chen.
\newblock Imface: A nonlinear 3d morphable face model with implicit neural representations.
\newblock In {\em Proceedings of the IEEE/CVF Conference on Computer Vision and Pattern Recognition}, pages 20343--20352, 2022.

\end{thebibliography}
}

\end{document}